\documentclass[11pt,a4paper]{article}
\usepackage{times,latexsym}
\usepackage{url}
\usepackage[T1]{fontenc}
\usepackage{graphicx}
\usepackage{amsmath}
\usepackage{threeparttable} 
\usepackage{amsthm}
\usepackage{amsmath}
\usepackage{amssymb}
\usepackage{booktabs}
\usepackage{multirow}
\usepackage{algorithm}
\usepackage{algorithmic}
\usepackage[switch]{lineno}
\usepackage{makecell}
\usepackage[edges]{forest}
\definecolor{hidden-draw}{RGB}{205, 44, 36}
\definecolor{hidden-blue}{RGB}{194,232,247}
\definecolor{hidden-orange}{RGB}{243,202,120}
\definecolor{hidden-yellow}{RGB}{242,244,193}
\definecolor{tree-level-1}{RGB}{245,20,85}
\definecolor{tree-level-2}{RGB}{246,86,118}
\definecolor{tree-level-3}{RGB}{248,177,193}
\definecolor{tree-leaf}{RGB}{176,230,198}

\usepackage{amsmath}

\newtheorem{remark}{Remark}
\usepackage[acceptedWithA]{tacl2021v1}
\usepackage{xspace,mfirstuc,tabulary}

\newif\iftaclinstructions
\taclinstructionsfalse % AUTHORS: do NOT set this to true
\iftaclinstructions
\renewcommand{\confidential}{}
\renewcommand{\anonsubtext}{(No author info supplied here, for consistency with
TACL-submission anonymization requirements)}
\newcommand{\instr}
\fi

\iftaclpubformat % this "if" is set by the choice of options

\else

\fi

\title{A Survey on Model Compression for Large Language Models}

\author{Xunyu Zhu$^{1,2}$, Jian Li$^{1,2}$\thanks{Corresponding author}, Yong Liu$^{3}$, Can Ma$^{1,2}$, Weiping Wang$^{1,2}$\\
$^1$Institute of Information Engineering, Chinese Academy of Sciences\\
$^2$School of Cyber Security, University of Chinese Academy of Sciences\\
$^3$Gaoling School of Artificial Intelligence, Renmin University of China\\
\texttt{\{zhuxunyu, lijian9026, macan, wangweiping\}@iie.ac.cn,
liuyonggsai@ruc.edu.cn}\\
}
\date{}

\begin{document}
\maketitle
\begin{abstract}
  Large Language Models (LLMs) have transformed natural language processing tasks successfully. Yet, their large size and high computational needs pose challenges for practical use, especially in resource-limited settings. Model compression has emerged as a key research area to address these challenges. This paper presents a survey of model compression techniques for LLMs. We cover methods like quantization, pruning, and knowledge distillation, highlighting recent advancements. We also discuss benchmarking strategies and evaluation metrics crucial for assessing compressed LLMs. This survey offers valuable insights for researchers and practitioners, aiming to enhance efficiency and real-world applicability of LLMs while laying a foundation for future advancements.	
\end{abstract}

\tikzstyle{my-box}=[
 rectangle,
 draw=hidden-draw,
 rounded corners,
 text opacity=1,
 minimum height=1.5em,
 minimum width=5em,
 inner sep=2pt,
 align=center,
 fill opacity=.5,
 ]
 \tikzstyle{leaf}=[my-box, minimum height=1.5em,
 fill=hidden-orange!60, text=black, align=left,font=\scriptsize,
 inner xsep=2pt,
 inner ysep=4pt,
 ]
 \begin{figure*}[!ht]
	\centering
	\resizebox{\textwidth}{!}{
		\begin{forest}
			forked edges,
			for tree={
				grow=east,
				reversed=true,
				anchor=base west,
				parent anchor=east,
				child anchor=west,
				base=left,
				font=\small,
				rectangle,
				draw=hidden-draw,
				rounded corners,
				align=left,
				minimum width=4em,
				edge+={darkgray, line width=1pt},
				s sep=3pt,
				inner xsep=2pt,
				inner ysep=3pt,
				ver/.style={rotate=90, child anchor=north, parent anchor=south, anchor=center},
			},
			where level=1{text width=5.3em,font=\scriptsize}{},
			where level=2{text width=6.3em,font=\scriptsize}{},
			where level=3{text width=6.6em,font=\scriptsize}{},
            where level=4{text width=6.6em,font=\scriptsize}{},
			[
			Model Compression for Large Language Models, ver
            [
			Quantization ($\S$\ref{sec:quantization})
			[
			Quantization-Aware \\ Training ($\S$\ref{sec:qat})
			[
			LLM-QAT~\cite{abs-2305-17888}{,}
            BitDistiller~\cite{abs-2402-10631}{,}
            OneBit~\cite{abs-2402-11295}
			, leaf, text width=41.3em
			]
			]
			[
			Post-Training \\ Quantization ($\S$\ref{sec:ptq})
			[
			Weight-Only \\ Quantization
            ($\S$\ref{sec:weight-only-quantization})
			[
			LUT-GEMM~\cite{park2024lutgemm}{,}
            GPTQ~\cite{frantar2023optq}{,}
            QuIP~\cite{chee2023quip}{,}\\
			AWQ~\cite{abs-2306-00978}{,}
            OWQ~\cite{LeeJKKP24}{,} 
			SpQR~\cite{dettmers2024spqr}{,}\\
            SqueezeLLM~\cite{abs-2306-07629} 
			, leaf, text width=33em
			]
			]
			[
			Weight-Activation \\ Quantization
            ($\S$\ref{sec:weight-activation-quantization})
			[
            ZeroQuant~\cite{YaoAZWLH22}{,}
			LLM.int8()~\cite{DettmersLBZ22}{,}  
			SmoothQuant~\cite{XiaoLSWDH23}{,}\\
			RPTQ~\cite{abs-2304-01089}{,}
			OliVe~\cite{0003THL00LG023}{,} 
			OS+~\cite{wei-etal-2023-outlier}{,}
            LLM-FP4~\cite{liu-etal-2023-llm}{,}\\
            OmniQuant~\cite{shao2024omniquant}
			, leaf, text width=33em
			]
			]
            [
            KV Cache \\ Quantization ($\S$\ref{sec:kvcq})
            [
            KVQuant~\cite{abs-2401-18079}{,}
            KIVI~\cite{abs-2402-02750}{,}
            WKVQuant~\cite{abs-2402-12065}
            , leaf, text width=33em
            ]
            ]
			]
			]
			[
			Pruning ($\S$\ref{sec:pruning})
			[
			Unstructured  Pruning 	\\($\S$\ref{sec:unstructured-pruning})	
			[
			SparseGPT~\cite{FrantarA23}{,}
			Wanda~\cite{sun2024a}{,}
            SAMSP~\cite{10445737}{,}
            DSnoT~\cite{zhang2024dynamic}{,}\\
            Flash-LLM~\cite{10.14778/3626292.3626303}
			, leaf, text width=41em
			]
			]
			[
			Structured  Pruning \\($\S$\ref{sec:structured-pruning})
			[
			LLM-Pruner~\cite{ma2023llmpruner}{,} 
            Shortened LLaMA~\cite{kim2024mefomo}{,}
            FLAP~\cite{AnZYTW24}{,}
            SliceGPT~\cite{ashkboos2024slicegpt}{,}\\
            Sheared LLaMA~\cite{xia2024sheared}
			, leaf, text width=41em
			]
			]
            [
            Semi-structured  \\Pruning
            ($\S$\ref{sec:semi-structured-pruning})
            [
            E-Sparse~\cite{abs-2310-15929}{,}
            SparseGPT~\cite{FrantarA23}{,}
			Wanda~\cite{sun2024a}            
            , leaf, text width=41em
            ]
            ]
			]
			[
			Knowledge \\ Distillation
            ($\S$\ref{sec:kd})
			[
			Black-box KD ($\S$\ref{sec:black-box-kd})
            [
			Chain-of-Thought \\ ($\S$\ref{sec:CoTD})
			[
			MT-COT~\cite{li2024explanations}{,}
            CoT Prompting~\cite{MagisterMAMS23}{,} 
			Fine-tune-CoT~\cite{HoSY23}{,} \\
            SSLM~\cite{pmlr-v202-fu23d}{,} 
			SCOTT~\cite{WangWLGYR23}{,} 
            Distilling Step-by-Step~\cite{HsiehLYNFRKLP23}{,} \\
			SOCRATIC CoT~\cite{ShridharSS23}{,} 
            PaD~\cite{abs-2305-13888}{,}
            DRA~\cite{WangHLWSZHWDSZ23}{,}\\
            TDIG~\cite{LiYFPSWW024}
			, leaf, text width=33em
			]
			]
            [
			In-Context Learning \\($\S$\ref{sec:ICLD})
			[
			In-context Learning Distillation~\cite{abs-2212-10670}{,}
            AICD~\cite{liu2024learning}
			, leaf, text width=33em
			]               
			]
            [
			Instruction Following \\($\S$\ref{sec:IFD})
			[
			Lion~\cite{jiang-etal-2023-lion}{,}
            LaMini-LM~\cite{wu-etal-2024-lamini}{,}
            SELF-INSTRUCT~\cite{wang-etal-2023-self-instruct}{,}\\
            Selective Reflection-Tuning~\cite{abs-2402-10110}
			, leaf, text width=33em
			]
			]
			]
            [
			White-box KD ($\S$\ref{sec:white-box-kd})
			[
			MINILLM~\cite{gu2024minillm}{,} 
			GKD~\cite{agarwal2024generalized}{,}
            TED~\cite{LiangZZHCZ23}
			, leaf, text width=41.2em
			]
			]
			]
            [
            Low-Rank \\Factorization ($\S$\ref{sec:lrf})
            [
            LPLR~\cite{SahaSP23}{,}
            ASVD~\cite{abs-2312-05821}{,}
            LSAER~\cite{sharma2024the}
            , leaf, text width=49.1em
            ]
            ]
			]
		\end{forest}
	}
	\caption{Taxonomy of Model Compression methods for Large Language Models.}
	\label{categorization_of_LLMs}
\end{figure*}
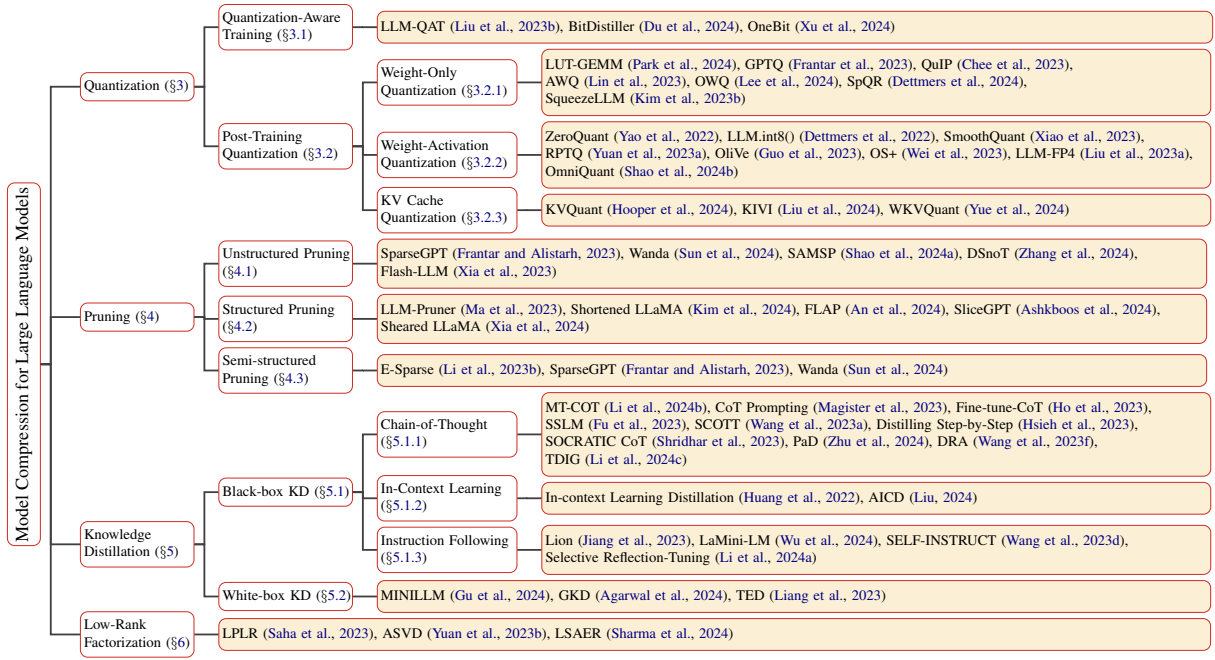

\section{Introduction}
Large Language Models (LLMs)~\cite{abs-2302-13971,abs-2307-09288,abs-2205-01068,abs-2211-05100,gpt-j,openai2024gpt4} refer to Transformer language models that contain billions (or more) of parameters, which are trained on massive text data. LLMs consistently exhibit remarkable performance across various tasks, but their exceptional capabilities come with significant challenges stemming from their extensive size and computational requirements. For instance, the GPT-175B model~\cite{BrownMRSKDNSSAA20}, with an impressive 175 billion parameters, demands a minimum of 350GB of memory in half-precision (FP16) format. Furthermore, deploying this model for inference necessitates at least five A100 GPUs, each featuring 80GB of memory, to efficiently manage operations. To tackle these issues, a prevalent approach known as model compression~\cite{HanMD15} offers a solution. Model compression involves transforming a large, resource-intensive model into a compact version suitable for deployment on resource-constrained devices. Additionally, model compression can enhance LLM inference speed and optimizes resource efficiency.

% Apart from their technical aspects, LLMs have triggered discussions on environmental and ethical matters. These models pose significant challenges for engineers and researchers in developing nations, where limited resources can impede access to essential hardware for model execution~\cite{abs-2306-00978}. Additionally, the substantial energy consumption of LLMs contributes to carbon emissions, underscoring the significance of sustainable practices in AI research. A promising solution to these challenges lies in utilizing model compression techniques, which have showcased the potential to reduce emissions without substantially compromising performance~\cite{abs-2211-02001}. By implementing model compression, we can tackle environmental concerns, enhance accessibility, and promote inclusivity in LLM deployment.

\begin{figure*}[]
    \centering
    \includegraphics[width=\textwidth]{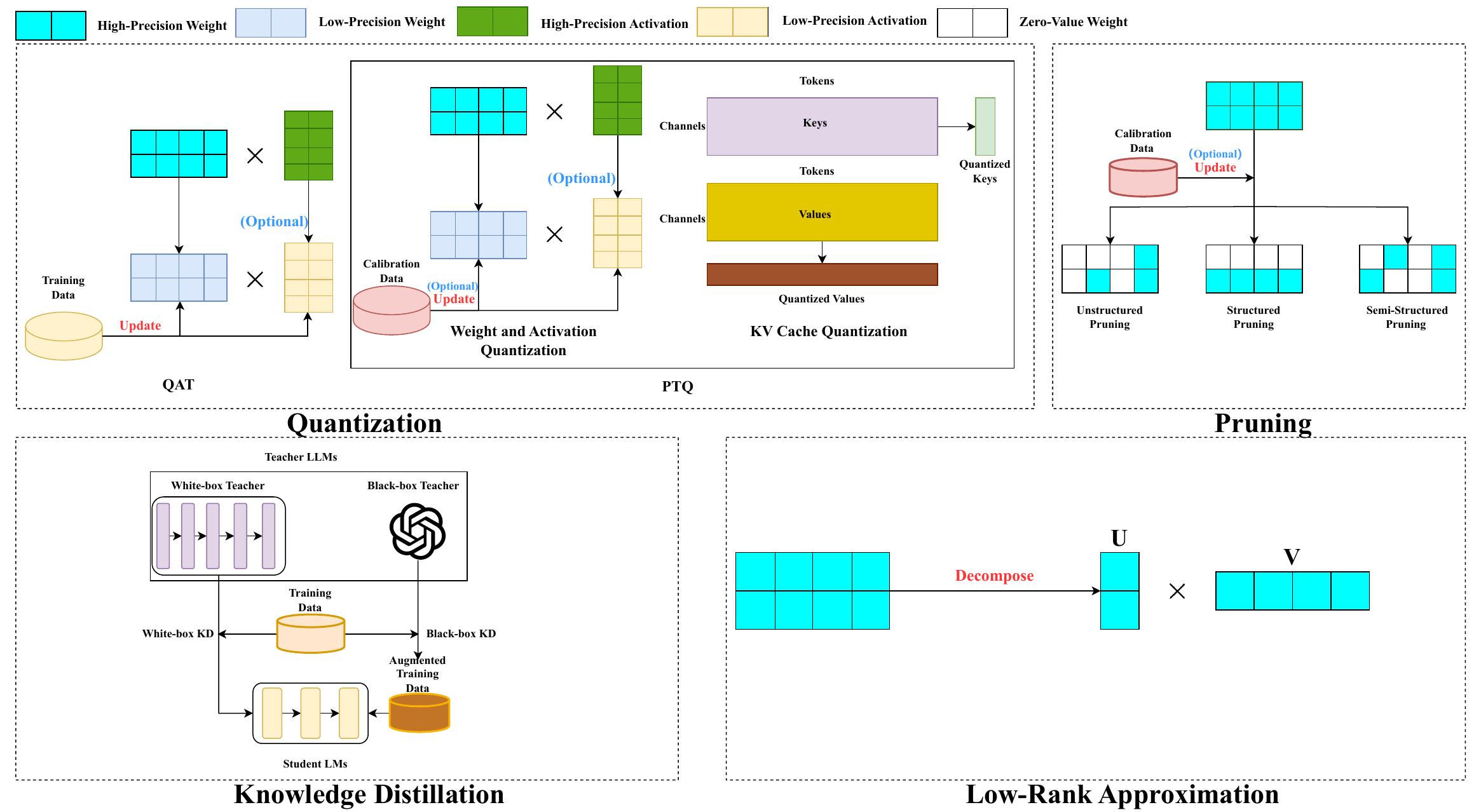}
    \caption{Illustrations of model compression methods for LLMs. In these methods, Quantization-Aware Training (QAT) and Knowledge Distillation (KD) stand out as task-based model compression techniques, tailored for specific tasks. Conversely, other model compression methods are task-agnostic, designed to operate independently of specific tasks. }
    \label{fig:model_compression_method}
\end{figure*}

In our paper, our primary objective is to illuminate the recent strides made in the domain of model compression techniques tailored specifically for LLMs. Our work conducts an exhaustive survey of methodologies, metrics, and benchmarks of model compression for LLMs. Figure~\ref{categorization_of_LLMs} shows the taxonomy of model compression methods for LLMs,  including quantization, pruning, knowledge distillation, and low-rank factorization. Figure~\ref{fig:model_compression_method} further shows basic flow of these  model compression methods for LLMs. Furthermore, our study sheds light on prevailing challenges and offers a glimpse into potential future research trajectories in this evolving field. We advocate for collaborative efforts within the community to pave the way for an ecologically conscious, all-encompassing, and sustainable future for LLMs. While there were previous surveys on neural networks model compression~\cite{LiLM23} and it has been lightly discussed in prior surveys on LMs~\cite{rogers-etal-2020-primer} and LLMs~\cite{abs-2303-18223}, our work is the inaugural survey dedicated solely to model compression for LLMs.

\section{Metrics and Benchmarks}
\subsection{Metrics}
Model compression of LLMs can be measured using various metrics, which capture different aspects of performance. These metrics are commonly presented alongside accuracy and zero-shot ability to comprehensively evaluate the LLM.
 
\textbf{Model Size} in a LLM typically  is measured by the number of total parameters of the LLM.  In general, LLMs with more parameters often requires more computational resources and memory for both training and inference.

\textbf{Floating Point Operations (FLOPs)} is an indicator that measures the  computational efficiency of LLMs, representing the number of floating-point operations required for the LLM to perform an instance.  In model compression, reducing FLOPs helps to make the LLM run faster and more efficiently. 

\textbf{Mean FLOPS Utilization (MFU)} quantifies the practical efficiency of computational resource utilization by LLMs during tasks.  MFU measures the ratio of actual FLOPS utilized by the LLM to the maximum theoretical FLOPS of a device. Unlike FLOPs, which estimates the maximum operations an LLM might perform, MFU assesses the actual effectiveness of resource use in operation. Essentially, while FLOPs measures a LLM's theoretical compute needs, MFU shows how effectively these computations are utilized in practice.

\textbf{Inference time} (i.e., latency) measures the time taken by the LLM to process and generate responses for input data during inference. Inference time is particularly crucial for real-world applications where the LLM needs to respond for user queries or process large amounts of data in real-time.
 
\textbf{Speedup Ratio} measures how much faster a compressed LLM performs tasks compared to the original LLM.  Specifically, it measures the ratio of the inference time of the uncompressed model over the inference time of the compressed model. Higher ratios mean greater efficiency and reduced computation time, highlighting effective compression. 

\textbf{Compression Ratio} measures how much a LLM's size is reduced through compression, calculated as the original size divided by the compressed size. Higher ratios mean greater size reduction, showing the compression's effectiveness in saving storage and memory.

\subsection{Benchmarks and Datasets} 
 The main goal of these benchmarks and datasets is to measure the efficiency and performance of compressed LLMs in comparison to their uncompressed counterparts. These benchmarks and datasets typically consist of diverse tasks and datasets that cover a range of natural language processing challenges. 
 
 \subsubsection{Common Benchmarks and Datasets}
 The majority of research evaluates compressed LLMs on well-established NLP benchmarks and datasets. For instance, WikiText-2~\cite{MerityX0S17}, C4~\cite{RaffelSRLNMZLL20}, and PTB~\cite{10.5555/972470.972475}  are designed for evaluating the perplexity performance of language models. LAMBADA~\cite{PapernoKLPBPBBF16}, PIQA~\cite{TataP03}, and OpenBookQA~\cite{MihaylovCKS18} are designed to evaluate the zero-shot ability of language models. GSM8K~\cite{abs-2110-14168}, CommonsenseQA~\cite{talmor-etal-2019-commonsenseqa} and StrategyQA~\cite{GevaKSKRB21} are designed to evaluate the reasoning  ability of language models.

\subsubsection{BIG-Bench}
BIG-Bench (BBH)~\cite{srivastava2023beyond}  is a benchmark suite designed for LLMs, covering over 200 NLP tasks, e.g., Text Comprehension Tasks, Inference Tasks, Mathematical Reasoning Tasks. The aim of BBH is to evaluate the performance of LLMs across these various complex tasks.  The compressed LLMs use BBH to measure their capability across a multidimensional spectrum of tasks. 

\subsubsection{Unseen Instructions Datasets}
Unseen instructions datasets are used to evaluate the performance of LLMs on unseen tasks. For instance, the Vicuna-Instructions~\cite{vicuna2023} dataset created by GPT-4 includes 80 complex questions across nine different categories like generic, knowledge-based, and writing tasks. Another dataset, User-Oriented-Instructions~\cite{WangKMLSKH23}, consists of 252 carefully selected instructions inspired by various user-focused applications such as Grammarly, StackOverflow, and Overleaf. These datasets evaluate how well compact LLMs can handle and carry out new tasks by presenting them with unfamiliar instructions.

\subsubsection{EleutherAI LM Harness} 
The EleutherAI LM Harness~\cite{eval-harness}  is an advanced framework for evaluating LLMs, providing a unified testing platform that supports over 60 standard academic benchmarks along with hundreds of subtasks and variants. The standardized evaluation tasks provided by the harness ensure the reproducibility and comparability of evaluation, which is essential for implementing fair and reproducible evaluations for the compressed LLMs.

\begin{table*}[]
\centering
\caption{The performance of various representative LLM quantization methods. }
\label{tab:quantization}
\resizebox{\textwidth}{!}{%
\begin{threeparttable}
\begin{tabular}{@{}ccccccccc@{}}
\toprule
\multirow{2}{*}{\textbf{Category}\tnote{$\dagger$}} & \multirow{2}{*}{\textbf{Methods}} & \multirow{2}{*}{\textbf{LLM}} & \multicolumn{3}{c}{\textbf{Bit Width}} & \multicolumn{2}{c}{\textbf{Perplexity Difference}\tnote{$\ddagger$}} & \multirow{2}{*}{\textbf{Speedup}} \\ \cmidrule(lr){4-8}
 &  &  & \textbf{Weights} & \textbf{Activations} & \textbf{KV Cache} & \textbf{Wikitext-2} & \textbf{C4} &  \\ \midrule
\multirow{3}{*}{QAT} & LLM-QAT & LLaMA-30B & 4 & 8 & 16 & 0.5 & 0.9 & - \\
 & BitDistiller & LLaMA2-13B & 2 & 16 & 16 & 1.9 & - & - \\
 & OneBit & LLaMA-13B & 1 & 16 & 16 & 4.09 & 3.64 & - \\ \midrule
\multirow{7}{*}{Weight-Only Quantization} & LUT-GEMM & LLaMA-65B & 3 & 16 & 16 & 0.14 & - & 2.04$\times$ \\
 & SqueezeLLM & LLaMA-13B & 3 & 16 & 16 & 0.51 & 0.67 & 2.4$\times$ \\
 & GPTQ & OPT-175B & 3 & 16 & 16 & 0.34 & 0.23 & 3.24$\times$ \\
 & AWQ & LLaMA2-70B & 3 & 16 & 16 & 0.42 & - & 3.2$\times$ \\
 & OWQ & LLaMA-65B & 3.01 & 16 & 16 & 0.72 & - & - \\
 & SpQR & LLaMA-30B & 3.89 & 16 & 16 & 0.15 & 0.1 & 2.0$\times$ \\
 & QuIP & LLaMA2-70B & 2 & 16 & 16 & 3.007 & 3.228 & - \\ \midrule
\multirow{9}{*}{Weight-Activation Quantization} & ZeroQuant & GPT-J-6B & 8 & 8 & 16 & 0.16 & - & 3.67$\times$ \\
 & LLM.int8() & OPT-13B & 8 & 8 & 16 & - & 0.00 & 1.22$\times$ \\
 & SmoothQuant & OPT-175B & 8 & 8 & 16 & 0.18 & - & 1.56$\times$ \\
 & RPTQ & OPT-175b & 4 & 4 & 16 & 2.26 & 2.15 & - \\
 & Olive & BLOOM-7B & 4 & 4 & 16 & 2.11 & 2.24 & 4.5$\times$ \\
 & OS+ & LLaMA-65B & 4 & 4 & 16 & 5.77 & - & - \\
 & QT & OPT-1.3B & 8 & 8 & 16 & 17.74 & - & - \\
 & ZeroQuant-FP & LLaMA-30B & 4 & 8 & 16 & 0.18 & 0.13 & - \\
 & OmniQuant & LLaMA-7B & 4 & 6 & 16 & 0.41 & 0.55 & - \\ \midrule
\multirow{2}{*}{KV Cache Quantization} & KVQuant & LLaMA-65B & 16 & 16 & 2 & 0.19 & 0.11 & 1.4$\times$ \\
 & WKVQuant & LLaMA-13B & 4 & 16 & 4 & 0.12 & 0.14 & - \\ \bottomrule
\end{tabular}
\begin{tablenotes} 
\item[$\dagger$]: The results presented in the table are solely derived from the original papers.
\item[$\ddagger$]: (The perplexity of the quantized LLM) - (The perplexity of the origin LLM).
\end{tablenotes}  
\end{threeparttable}
}
\end{table*}

\section{Quantization}
\label{sec:quantization}
Quantization~\cite{720541} refers to the process of reducing the number of bits (i.e., precision) in the parameters of the model with minimal loss in inference performance. Quantization can be categorized into two main approaches: \textbf{Quantization-Aware Training (QAT)}, and \textbf{Post-Training Quantization (PTQ)}. The primary distinction between the two approaches lies in whether retraining is needed during quantization. PTQ enables direct use of quantized models in inference, while QAT requires retraining to rectify errors introduced by quantization. Table~\ref{tab:quantization} shows the performance of many representative LLM quantization methods.

\subsection{Quantization-Aware Training}
\label{sec:qat}
QAT involves retraining a quantized model to counteract performance degradation caused by quantization.  For instance, LLM-QAT~\cite{abs-2305-17888} implements the standard QAT framework directly onto LLMs.  LLM-QAT distills knowledge by generating data from the LLM itself, and train the quantized LLM to align with the output distribution of the original LLM based on the generated data.   BitDistiller~\cite{abs-2402-10631} merges QAT with self-distillation, enhancing LLM performance at sub-4-bit precisions. It employs tailored asymmetric quantization, clipping, and a Confidence-Aware Kullback-Leibler Divergence objective for faster convergence and superior results.  OneBit~\cite{abs-2402-11295} introduces a novel 1-bit parameter representation method and an effective parameter initialization method to implement 1-bit quantization for LLM weight matrices, paving the way for the extremely low bit-width deployment of LLMs. 

\begin{remark}
While QAT can mitigate quantization's accuracy degradation, retraining demands  a lot of effort due to tens or hundreds of billions of parameters in LLMs. A practical solution is to incorporate Parameter-Efficient Fine-Tuning (PEFT) into the retraining process of QAT. Currently, methods like QLORA~\cite{DettmersPHZ23}, PEQA~\cite{abs-2305-14152} and LoftQ~\cite{abs-2310-08659} combine quantization with PEFT for model fine-tuning efficiency. However, these methods are typically task-dependent. L4Q~\cite{abs-2402-04902} makes a preliminary attempt to enhance generality by leveraging LoRA-wise learned quantization step size for LLMs. We think that introducing PEFT to enhance QAT efficiency is not only feasible but also holds significant promise, warranting thorough exploration.
\end{remark}

\subsection{Post-Training Quantization}
\label{sec:ptq}
 PTQ efficiently converts a full-precision LLM to low-precision without retraining, saving memory and computational costs. We categorize PTQ for LLMs into three groups: \textbf{Weight-Only Quantization}, \textbf{Weight-Activation Quantization}, and \textbf{KV Cache Quantization}. The disparity between these groups lies in their quantization objectives. Weight-only quantization focuses solely on quantizing weights, whereas weight-activation quantization extends its objective to both weights and activations.  Prior research~\cite{abs-2303-08302} indicates that activation quantization is typically more sensitive to weight quantization, allowing weight-only quantization to achieve lower bit-width. However, since quantized weights necessitate dequantization before multiplication with activations, weight-only quantization inevitably introduces additional computational overhead during inference and  cannot enjoy the accelerated low-bit operation supported by specific hardware. Furthermore,  kv cache quantization targets the KV cache, which stores keys and values of attention layers. The KV cache often consumes lots of memory, acting as a bottleneck for input streams containing lengthy tokens.  By implementing kv cache quantization, it is possible to increase throughput and accommodate inputs with longer tokens more efficiently.

\subsubsection{Weight-Only Quantization}
\label{sec:weight-only-quantization}
Weight-only quantization is the most conventional and widespread method. For example, LUT-GEMM~\cite{park2024lutgemm} uses binary-coding quantization (BCQ)~\cite{RastegariORF16} format, which factorizes the parameters of LLMs into binary parameters and a set of scaling factors, to accelerate quantized matrix multiplications in weight-only quantization. GPTQ~\cite{frantar2023optq} proposes a layer-wise quantization method based on Optimal Brain Quantization (OBQ)~\cite{frantar2022optimal}, which updates weights with inverse Hessian information, and quantizes LLMs into 3/4-bit. QuIP~\cite{chee2023quip}  optimally adjusts weights by utilizing the LDL decomposition of the Hessian matrix derived from vectors drawn uniformly at random from a calibration set, and multiplies weight and Hessian matrices with a Kronecker product of random orthogonal matrices to ensure incoherence between weight and Hessian matrices. Combining these two steps, QuIP successfully quantizes LLMs into 2-bits with minimal performance loss.

To further minimize quantization errors in the weight-only quantization of LLMs, lots of works identify sensitive weights, which have an important effect on LLMs' quantization performance, and store these sensitive weights in high precision. For example, AWQ~\cite{abs-2306-00978} stores the top 1\% of weights that have the most significant impact on LLM performance in high-precision, and integrates a per-channel scaling method to identify optimal scaling factors. Here, "channel" denotes individual dimensions or feature maps within the model. Similar with AWQ, OWQ~\cite{LeeJKKP24} store weights sensitive to activation outliers in high-precision, and quantizes other non-sensitive weights.  Different from OWQ, SpQR~\cite{dettmers2024spqr} employs the L2 error between the original and quantized predictions as a weight sensitivity metric.  Furthermore, SqueezeLLM~\cite{abs-2306-07629} introduces a  weights clusters algorithm  based on sensitivity, using k-means centroids as quantized weight values, to identify sensitive weights. The sensitivity is approximated by the Hessian matrix of weights.  Then, SqueezeLLM stores sensitive weights in an efficient sparse format, and quantize other weights. SqueezeLLM quantizes LLMs in 3-bit, and achieves a more than 2× speedup compared to the FP16 baseline.

\subsubsection{Weight-Activation Quantization} 
\label{sec:weight-activation-quantization}
Alongside works centered on weight-only quantization in LLMs, there is a plethora of research focusing primarily on weight-activation quantization in LLMs. For example, ZeroQuant~\cite{YaoAZWLH22} is the first work to implement weight-activation quantization for LLMs, which uses group-wise quantization for weight and token-wise quantization for activations, and  reduces the precision for weights and activations of LLMs to INT8. 

LLMs have outliers in activations, and the performance of LLMs declines a lot, if these activations with outliers are directly quantized. Recent works try to treat these outliers specially to reduce quantization errors in weight-activation quantization. For example, LLM.int8()~\cite{DettmersLBZ22} stores these outlier feature dimensions into high-precision, and uses vector-wise quantization, which assigns separate normalization constants to each inner product within matrix multiplication, to quantize other features. LLM.int8() quantizes weights and activations of LLMs into 8-bit without any performance degradation. SmoothQuant~\cite{XiaoLSWDH23} designs a per-channel scaling transformation to smooths the activation outliers based on the discovery that different tokens have similar variations across channels of activations. RPTQ~\cite{abs-2304-01089} finds that the range of values varies greatly between different channels, and integrates a channel reordering method, which clusters and reorders the channels in the activation and uses the same quantization parameters to quantize the values in each cluster, into layer normalization and linear layer weights to efficiently reduce the effect of numerical range differences between channels. OliVe~\cite{0003THL00LG023} thinks that outliers are more important than the normal values, and uses an outlier-victim pair
(OVP) quantization to handle outlier values locally with low hardware overheads and
significant performance benefits. OS+~\cite{wei-etal-2023-outlier} further finds that outliers are concentrated in specific and asymmetric channels. Based on the findings, OS+ incorporates channel-wise shifting to eliminate the impact of asymmetry and channel-wise scaling to balance the distribution of outliers. LLM-FP4~\cite{liu-etal-2023-llm} uses floating-point formats (specifically FP8 and FP4) to address the limitations of traditional integer quantization (such as INT8 and INT4) to deal with outliers. Furthermore, LLM-FP4~\cite{liu-etal-2023-llm} points out that exponent bits and clipping range are important factors that effect the performance of FP quantization, and introduces a search-based framework for determining the optimal exponent bias and maximal quantization value. OmniQuant~\cite{shao2024omniquant} handles the activation outliers by equivalently shifting the challenge of quantization from activations to weights, and optimizes the clipping threshold to adjust the extreme values of the weights.

\subsubsection{KV Cache Quantization}
\label{sec:kvcq}
With the increasing number of input tokens supported by LLMs, the memory usage of the KV cache also increases. Recent efforts begin to focus on kv cache quantization to reduce the memory footprint of LLMs and accelerate their inference. For example, KVQuant~\cite{abs-2401-18079} proposes several KV Cache
Quantization methods, such as Per-Channel Key Quantization, PreRoPE Key Quantization, and Non-Uniform kv cache quantization, to implement
10 million context length LLM inference.   Through an in-depth analysis of the element distribution within the KV cache, KIVI~\cite{abs-2402-02750} finds that key caches should be quantized per-channel, while value caches should be quantized per-token. Finally, KIVI succeeds in quantizing the KV cache to 2 bits without fine-tuning.  WKVQuant~\cite{abs-2402-12065} presents an innovative approach for quantizing large language models (LLMs) by integrating past-only quantization to refine attention computations, employing a two-dimensional quantization strategy to manage the distribution of key/value (KV) caches effectively, and utilizing cross-block reconstruction regularization for optimizing parameters. This method enables the quantization of both weights and KV caches, resulting in memory savings that rival those of weight-activation quantization, while nearly matching the performance levels of weight-only quantization.

% \begin{remark}
%      \textcolor{red}{For PTQ, preparing a high-quality calibration dataset to assist in determining the right quantization weights and scaling factors is crucial. Specifically, ~\citet{abs-2311-09755} make a extensive empirical study on the effect of calibration data upon  model compression methods, and  find that the performance of downstream tasks can vary significantly depending on the calibration data selected. High-quality calibration data can improve the performance and accuracy of the quantized model, so careful selection and preparation of calibration data are necessary.}
% \end{remark}

\begin{table*}[]  
\centering  
\scriptsize  
\begin{threeparttable} 
\caption{The performance of various representative LLM pruning methods.} 
\label{tab:pruning} 
\begin{tabular}{@{}cccccc@{}}  
\toprule  
\multirow{2}{*}{\textbf{Category}\tnote{$\dagger$}} & \multirow{2}{*}{\textbf{Methods}} & \multirow{2}{*}{\textbf{LLM}} & \multirow{2}{*}{\begin{tabular}[c]{@{}c@{}}\textbf{Perplexity Difference}\\ \textbf{(WikiText-2)}\tnote{$\ddagger$}\end{tabular}} & \multirow{2}{*}{\textbf{Compression Rate}} & \multirow{2}{*}{\textbf{Speed up}} \\ 
 &  &  &&  &  \\ \midrule  
\multirow{4}{*}{Unstructured Pruning} & SparseGPT & OPT-175B & -0.14 & 50\% & - \\  
 & Wanda & LLaMA-65B & 1.01 & 50\% & - \\  
 & SAMSP & LLaMA2-13B & 0.63 & 50\% & - \\  
 & DSnoT & LLaMA-65B & 2.08e4 & 90\% & - \\ \midrule  
\multirow{4}{*}{Structured Pruning} & LLM-Pruner & LLaMA-13B & 3.6 & 20\% & - \\  
 & Shortened LLaMA & LLaMA-7B & 10.5 & 35\% & - \\  
 & FLAP & LLaMA-65B & 7.09 & 50\% & - \\  
 & SliceGPT & LLaMA2-70B & 1.73 & 30\% & 1.87$\times$ \\ \midrule  
\multirow{3}{*}{Semi-Structured Pruning} & E-Sparse & LLaMA-65B & 2.13 & 2:4 & 1.53$\times$ \\  
 & SparseGPT & OPT-175B & 0.39 & 2:4 & 2$\times$ \\  
 & Wanda & LLaMA-65B & 2.69 & 2:4 & 1.24$\times$ \\ \bottomrule  
\end{tabular}  
\begin{tablenotes} 
\item[$\dagger$]: The results presented in the table are solely derived from the original papers.
\item[$\ddagger$]: (The perplexity of the pruned LLM) - (The perplexity of the origin LLM).
\end{tablenotes}  
\end{threeparttable}  
\end{table*}  

\section{Pruning}
\label{sec:pruning}
Pruning~\cite{CunDS89} is a powerful technique to reduce the size or complexity of a model by removing redundant components.   Pruning can be divided into \textbf{Unstructured Pruning}, \textbf{Semi-Structured Pruning}, and \textbf{Structured Pruning}. Structured pruning removes entire components like neurons, attention heads, or layers based on specific rules while preserving the overall network structure. On the other hand, unstructured pruning prunes individual parameters, resulting in an irregular sparse structure.  Semi-structured pruning is a method that lies between structured pruning and unstructured pruning, capable of achieving fine-grained pruning and structural regularization simultaneously. It prunes partial parameters based on specific patterns rather than entire channels, filters, or neurons, making it a fine-grained form of structured pruning. Table~\ref{tab:pruning} shows the performance of many representative LLM pruning methods.

\subsection{Unstructured Pruning}
\label{sec:unstructured-pruning}
Unstructured pruning preserves the pruned model's performance, hence, works related to unstructured pruning of LLMs often dispense with retraining to restore performance. Nevertheless, unstructured pruning renders the pruned model irregular, necessitating specialized handling or software optimizations for inference acceleration.
An innovative approach in this domain is SparseGPT~\cite{FrantarA23}, which introduces a one-shot pruning strategy without retraining. SparseGPT frames pruning as an extensive sparse regression problem and solves it using an approximate sparse regression solver. SparseGPT achieves significant unstructured sparsity, even up to over 50\% on the largest GPT models like OPT-175B and BLOOM-176B, with minimal increase in perplexity. To reduce the cost about the weight update process required by SparseGPT, Wanda~\cite{sun2024a}  achieves model sparsity by pruning weights with the smallest magnitudes multiplied by the norm of  the corresponding input activations, without the need for retraining or weight updates. To further minimize pruning-induced errors while upholding the desired overall sparsity level, SAMSP~\cite{10445737} utilizes the Hessian matrix as a metric for weight matrix sensitivity evaluation, and dynamically adjusts sparsity allocation based on sensitivity. Furthermore, DSnoT~\cite{zhang2024dynamic} minimizes the reconstruction error between dense and sparse models through iterative weight pruning-and-growing on top of sparse LLMs to enhance LLM performance across various sparsity rates, especially at high sparsity levels. 
To provide hardware support for handling unstructured  pruning on  the GPU Tensor Core hardware, Flash-LLM~\cite{10.14778/3626292.3626303} introduces an unstructured sparse matrix multiplication method, which loads weight matrices in a sparse format from global memory and reconstructs them in a dense format within high-speed on-chip buffers for computation using tensor cores. 

 % \begin{remark}
 %   \begin{enumerate}
 %       \item \textcolor{red}{Choosing the optimal pruning strategy is crucial for maximizing performance and compatibility with the target hardware. For instance, SparseGPT utilizes the DeepSparse engine~\cite{deepsparse} to evaluate the OPT-2.7B model, demonstrating that unstructured pruning yields close-to-optimal acceleration on CPUs. Moreover, conducting a 2:4 sparsity test on various layers of the OPT-175B model with the CUTLASS library reveals notable speedups ranging from 54\% to 79\% on GPUs, underscoring the significant performance enhancement achievable with semi-structured pruning on GPU architectures.} 
 %       \item \textcolor{red}{When language models are pruned, the performance of larger models declined less than that of their smaller counterparts. For example, by pruning OPT-175B with SparseGPT to achieve a 50\% sparsity, the model's perplexity decreased by 0.14\% compared to the original OPT-175B on raw-WikiText2. On the other hand, pruning OPT-2.7B with SparseGPT to achieve a 50\% sparsity resulted in a 1.01\% increase in perplexity compared to the original OPT-175B on raw-WikiText2. Thus, efficient pruning strategies are very important for large language models.}
 %   \end{enumerate}
 % \end{remark} 

\subsection{Structured Pruning}
\label{sec:structured-pruning}
Compared to unstructured pruning, structured pruning offers the advantage of being hardware-agnostic, allowing for accelerated inference on traditional hardware post-pruning. However, the removal of larger and potentially more critical components in structured pruning may result in performance degradation, typically requiring efficient parameter fine-tuning for recovery. We divide LLMs structured pruning works  into several groups based on  pruning metrics: \textbf{Loss-based Pruning}, \textbf{Magnitude-based Pruning}, \textbf{Regularization-based Pruning}. 
 
 \textbf{Loss-based Pruning}~\cite{MolchanovMTFK19} assesses the significance of a pruning unit by measuring its impact on loss or gradient information (e.g., first-order or second-order derivatives of loss). For example, LLM-Pruner~\cite{ma2023llmpruner} introduces a one-shot structured pruning on LLMs based on gradient information. Specifically, LLM-Pruner identifies dependent structures via a dependency detection algorithm and selects optimal pruning groups using gradient information, rather than solely relying on loss changes, in a task-agnostic manner.  Different from LLM-Pruner, which focuses on narrowing LLMs' width,  Shortened LLaMA~\cite{kim2024mefomo} introduces a one-shot depth pruning on LLMs. Shortened LLaMA chooses the Transformer block as the prunable unit, and prunes these unimportant Transformer blocks, where the importance of Transformer blocks is evaluated by loss and its second-order derivative. After Pruning, both LLM-Pruner and Shortened LLaMA utilize LoRA to rapidly recover the performance of the pruned model.

\textbf{Magnitude-based Pruning}~\cite{NIPS2015_ae0eb3ee}  involves devising a heuristic metric based on the magnitudes of pruning units, and use the metric to assess the importance of pruning units, subsequently pruning those units whose scores fall below a predefined threshold. For example, FLAP~\cite{AnZYTW24} utilizes a structured fluctuation metric to assess and identify columns in the weight matrix suitable for pruning, measuring the variation of each input feature relative to a baseline value to estimate the impact of removing a column of weights. Additionally, FLAP  uses an adaptive structure search to optimize global model compression, and restores the model's performance post-pruning through a baseline bias compensation mechanism, avoiding the need for fine-tuning. To further maintain the pruned model's performance,  SliceGPT~\cite{ashkboos2024slicegpt} leverages the computational invariance of transformer networks and optimizes the pruning process through Principal Component Analysis (PCA). Specifically, SliceGPT employs PCA as the pruning metric, applying it at each layer of the transformer network to project the signal matrix onto its principal components and eliminate insignificant columns or rows from the transformed weight matrices, ultimately aiming to compress the model effectively.

\textbf{Regularization-based Pruning}~\cite{NIPS2016_41bfd20a} typically adds a regularization term (e.g., $L_0$, $L_1$, and $L_2$ regularization) into the loss function to induce sparsity for LLMs. For example, Sheared LLaMA~\cite{xia2024sheared} uses a pair of Lagrange multipliers based on pruning masks to impose constraints on the pruned model shape directly, thereby formulating pruning as a constrained optimization problem.  Through solving this optimization problem, Sheared LLaMA derives optimal pruning masks. Additionally, Sheared LLaMA introduces dynamic batch loading, a strategy that adapts training data loading based on each domain's loss reduction rate, enhancing the efficiency of data utilization during training.

\begin{remark}
   Structured pruning typically reduces model size by removing redundant parameters, but it may degrade model performance. A novel approach is to combine knowledge distillation~\cite{HintonVD15} with structured pruning. Knowledge distillation allows knowledge extracted from a LLM to be transferred to a smaller model, helping the smaller model maintain its performance while reducing its size.
\end{remark}

\subsection{Semi-Structured Pruning}
\label{sec:semi-structured-pruning}
Apart from unstructured pruning and structured pruning, there are many works which use semi-structured pruning to prune partial weights  of LLMs based on specific patterns. N:M sparsity, where every M contiguous elements leave N non-zero elements, is an example of semi-structured pruning. For example, E-Sparse~\cite{abs-2310-15929} implements N:M sparsity by introducing information entropy as a metric for evaluating parameter importance to enhances the significance of parameter weights and input feature norms. E-Sparse incorporates global naive shuffle and local block shuffle to efficiently optimize information distribution and mitigate the impact of N:M sparsity on LLM accuracy. Furthermore, many pruning works can also be generalized to semi-structured patterns. For example, SparseGPT~\cite{FrantarA23} and Wanda~\cite{sun2024a} also explore N:M sparsity of LLMs. SparseGPT~\cite{FrantarA23} employs block-wise weight partitioning, with each block containing M weights. It identifies and prunes N weights with the lowest reconstruction error(based on Hessian information), ensuring a sparsity ratio of N:M. This process iteratively prunes and updates model weights, addressing one block at a time until the desired sparsity level is achieved across the entire model. Wanda~\cite{sun2024a} achieves structured N:M pruning by dividing the weight matrix into groups of M consecutive weights and computing an importance score for each weight. The score is determined by the product of the weight's magnitude and the norm of the corresponding input activations. Within each weight group, the N weights with the highest scores are retained, while the rest are set to zero, thereby implementing structured N:M pruning. Furthermore, choosing the optimal pruning strategy is crucial for compatibility with the target hardware. For instance, ~\citet{ChoquetteGGSK21}  introduce the Ampere Tensor Core GPU architecture (e.g., A100 GPUs) and propose 2:4 fine-grained semi-structured sparsity to accelerate Sparse Neural Networks on this hardware. However, the current implementation of the Ampere architecture supports only the 2:4 ratio, leaving other ratios without acceleration.

\begin{remark}
   LLMs often perform well on multiple tasks, which means they contain a multitude of parameters for various tasks. Dynamic  pruning~\cite{XiaWD20} methods can dynamically prune different parts of the model based on the current task's requirements to provide better performance on specific tasks. This helps strike a balance between performance and efficiency.
\end{remark}
\begin{remark}
For PTQ and pruning, preparing a high-quality calibration dataset to assist in improving the performance of compressed LLMs is crucial. Specifically, ~\citet{abs-2311-09755} make a extensive empirical study on the effect of calibration data upon  model compression methods, and  find that the performance of downstream tasks can vary significantly depending on the calibration data selected. High-quality calibration data can improve the performance and accuracy of the compressed model, so careful selection and preparation of calibration data are necessary.    
\end{remark}

\section{Knowledge  Distillation}
\label{sec:kd}
Knowledge Distillation (KD)~\cite{HintonVD15} is a  technique aimed at transferring knowledge from a large and complex model (i.e., teacher model) to a smaller and simpler model (i.e., student model).  We classify these methods into two clear categories~\cite{gu2024minillm}: \textbf{Black-box KD}, where only the teacher's outputs are accessible, typically from closed-source LLMs, and \textbf{White-box KD}, where the teacher's parameters or output distribution are available, usually from open-source LLMs.

% Due to the significant gap in capacity between student and teacher models, and considering that the  capabilities of teacher LLMs are distributed across a wide range of tasks, prior research~\cite{pmlr-v202-fu23d} has commonly attempted to distill specific abilities into student models to enhance their performance on particular tasks.

\subsection{Black-box KD}
\label{sec:black-box-kd}
Black-box KD usually prompts the teacher LLM to generate a distillation dataset for fine-tune the student LM, thereby transfering capabilities from teacher LLM to the student LM.
In Black-box KD, teacher LLMs such as ChatGPT (gpt-3.5-turbo) and GPT4~\cite{openai2024gpt4} are typically employed, while smaller LMs (SLMs), such as GPT-2~\cite{radford2019language}, T5~\cite{RaffelSRLNMZLL20}, FlanT5~\cite{abs-2210-11416}, and CodeT5~\cite{0034WJH21}, are commonly utilized as student LMs.  On the other hand, researchers find that LLMs have emergent abilities, which refers to a significant improvement in performance when the model reaches a certain scale, showcasing surprising capabilities. Lots of Black-box KD methods try to distill emergent abilities from LLMs to student LMs, and we introduce three commonly used emergent ability distillation methods: Chain-of-Thought (CoT) Distillation, In-Context Learning (ICL) Distillation,  and Instruction Following (IF) Distillation.

\subsubsection{Chain-of-Thought Distillation} 
\label{sec:CoTD}
CoT~\cite{Wei0SBIXCLZ22,0002WSLCNCZ23} prompts LLMs to generate intermediate reasoning steps, enabling them to tackle complex reasoning tasks step by step. ~\citet{li2024explanations} and ~\citet{HsiehLYNFRKLP23} employ LLMs to prompt the generation of explanations and leverage a multi-task learning framework to bolster the reasoning capabilities of smaller models while enhancing their capacity for generating explanations.   
~\citet{MagisterMAMS23} show that LLMs' reasoning capability can be transferred to SLMs via knowledge distillation, but there's a trade-off between model and dataset size in reasoning ability.  ~\citet{HoSY23} use zero-shot CoT techniques to prompt LLMs to generate diverse rationales to enrich the distillation dataset for the student models.    ~\citet{ShridharSS23} distill two student models: a problem decomposer and a subproblem solver, which the problem decomposer decomposes complex problems into a sequence of subproblems, and the subproblem solver solves these subproblems step by step. 
~\citet{WangWLGYR23} incorporate contrastive decoding during rationale generation for teacher models and address shortcut issues by introducing a counterfactual reasoning objective during student model training. ~\citet{pmlr-v202-fu23d} demonstrate that increasing task-specific capabilities through distillation may inadvertently lead to reduced performance in solving generalized problems, and focus on improving mathematical capability of student LMs via distillation. PaD~\cite{abs-2305-13888} prompts LLMs to generate Program-of-Thought (PoT) rationales instead of Chain-of-Thought (CoT) rationales to construct distillation dataset, and fine-tunes SLMs with the distillation dataset. ~\citet{WangHLWSZHWDSZ23} establishes a multi-round interactive learning paradigm that enables student LMs to provide feedback to teacher LLMs during the distillation process, thereby obtaining tailored training data. Additionally, DRA introduces a self-reflection learning mechanism, allowing the student LMs to learn from their mistakes and enhance their reasoning abilities. ~\citet{LiYFPSWW024} finds that negative data generated from teacher LMs also has reasoning knowledge, and  guides student LMs to learn knowledge from both negative samples besides positive ones.

\subsubsection{In-Context Learning Distillation}
\label{sec:ICLD}
ICL~\cite{abs-2301-00234,abs-2301-11916} employs structured prompts with task descriptions and examples for LLMs to learn new tasks without gradient updates. ~\citet{abs-2212-10670}  introduce a method called in-context learning distillation, which transfers in-context learning ability from LLMs to smaller models by combining in-context learning objectives with language modeling objectives. Specifically, it trains the student model to improve its generalization across various tasks by imitating the soft label predictions of the teacher model and the hard label ground truth values. Additionally, the method incorporates two few-shot learning paradigms: Meta In-context Tuning (Meta-ICT) and Multitask In-context Tuning (Multitask-ICT).   In Meta-ICT, the student model adapts to new tasks with in-context learning and guidance from the teacher. Conversely, Multitask-ICT treats all target tasks as training tasks, directly using examples from them in distillation. The outcomes show that Multitask-ICT is more effective, despite its increased computational requirements. AICD~\cite{liu2024learning}  leverages the autoregressive nature of LLMs to perform meta-teacher forcing on CoTs within the context, jointly optimizing the likelihood of all in-context CoTs, thereby distilling the capabilities of in-context learning and reasoning into smaller models.

\subsubsection{Instruction Following Distillation}
\label{sec:IFD}
IF~\cite{Ouyang0JAWMZASR22,brooks2023instructpix2pix} aims to bolster the zero-shot ability of LLMs through fine-tuning using a collection of instruction-like prompt-response pairs. For instance, Lion~\cite{jiang-etal-2023-lion}  prompts the LLM to identify and generate the "hard" instructions, which are then utilized to enhance the student model's capabilities. LaMini-LM~\cite{wu-etal-2024-lamini}  develops an extensive collection of 2.58 million instructions, comprising both existing and newly generated instructions, and fine-tunes a diverse array of models by using these instructions. SELF-INSTRUCT~\cite{wang-etal-2023-self-instruct} uses student LMs themselves as teachers to generate instruction following dataset, and fine-tunes students themselves with the dataset. Selective Reflection-Tuning~\cite{abs-2402-10110} leverages the teacher LLMs to reflect on and improve existing data, while the student LMs assess and selectively incorporate these improvements, thereby increasing data quality and compatibility with the student LMs. 

\begin{remark}
   Black-Box Distillation uses the teacher model's outputs as supervision, but the teacher model's outputs may not cover all possible input scenarios. Thus, understanding how to handle a student model's generalization on unknown data and how to increase data diversity is an area that requires further investigation.
\end{remark}

% \item  \textcolor{red}{Black-Box Distillation transfers knowledge from the teacher model, but gaining a deeper understanding of the knowledge transfer process may help improve the method. In LLMs, designing new prompting strategy is a feasible scheme. For example, ~\citet{ShridharSS23} design a prompt to instruct LLMs to decouple a complex problem into several simple problems, and then train two distilled models: a problem decomposer and a subproblem solver to improve the performance of distillation.}

\subsection{White-box KD}
\label{sec:white-box-kd}
White-box KD enables the student LM to gain a deeper understanding of the teacher LLM's internal structure and knowledge representations, often resulting in higher-level performance improvements.  An representative example is MINILLM~\cite{gu2024minillm}, which the first work to study distillation from the Open-source generative LLMs. MINILLM use a reverse Kullback-Leibler divergence objective, which is more suitable for KD on generative language models, to prevent the student model from overestimating the low-probability regions of the teacher distribution, and derives an effective optimization approach to learn the objective. Further, GKD~\cite{agarwal2024generalized} explores distillation from auto-regressive models, where generative language models are a subset. GKD  trains the student using self-generated outputs, incorporating teacher feedback, and allows flexibility in using different loss functions when the student cannot fully replicate the teacher's distribution. Different from the above works, which focus on learning the teacher distribution, TED~\cite{LiangZZHCZ23} proposes a task-aware layer-wise distillation method, which designs task-aware filters, which align the hidden representations of the teacher and student models at each intermediate layer, to reduce the knowledge gap between the student and teacher models.

\begin{remark}
  Although white-box distillation allows student LMs to learn the knowledge of teacher LLMs more deeply compared to black-box distillation, currently, open-source LLMs perform worse than closed-source ones, limiting the improvement of student LMs performance in white-box distillation. This is one of the barren factors hindering the development of white-box distillation. A feasible solution is to distill knowledge from closed-source LLMs to open-source LLMs through black-box distillation, and then use white-box distillation to transfer knowledge from open-source LLMs to student LLMs.
\end{remark}
\begin{remark}
White-box distillation often involves understanding and utilizing the internal structure of LLMs, such as layer connections and parameter settings. A more in-depth exploration of different network structures and interactions between layers can improve the effectiveness of white-box distillation.    
\end{remark}

\section{Low-Rank Factorization}
\label{sec:lrf}
Low-Rank Factorization~\cite{SrebroJ03} reduces a large matrix into smaller ones to save space and computational effort.  For example, it decomposes a large matrix $W$ into two small matrices $U$ and $V$ (i.e., $W \approx UV$), where $U$ is $m \times k$ and $V$ is $k \times n$, with $k$ much smaller than $m$ and $n$.  Recent works try to employ low-rank factorization to compress LLMs and achieve significant success in this regard. For example, LPLR~\cite{SahaSP23} compresses weight matrices of LLMs through randomized low-rank and low-precision factorization. Specifically, LPLR approximates the column space of the matrix using random sketching techniques, quantizes these columns, and then projects the original columns onto this quantized space to create two low-rank factors stored in low-precision.  ASVD~\cite{abs-2312-05821} finds that the activation distribution has an effect on the compression performance. To sovle the problem, ASVD proposes to scale the weight matrix with a diagonal matrix that contains scaling factors corresponding to the activation distribution of the input feature channels. Moreover, ASVD assigns the most suitable compression ratio to different layers by analyzing the singular values distribution in each layer's weight matrix, ensuring minimal loss of model performance during the compression process.   Furthermore,~\citet{sharma2024the} demonstrates that the performance of LLMs can be significantly improved by applying Layer-Selective Rank Reduction (LASER) to specific layers of Transformer models. LASER involves selectively reducing the rank higher-order components of weight matrices, which is shown to improve the model's handling of rare training data and its resistance to question paraphrasing.

% \begin{figure}[ht]
%  \centering
%  \includegraphics[width=\columnwidth]{kd.pdf}
%  \caption{A brief classification of knowledge distillation for LLMs.}
%  \label{kd}
%  \end{figure}

\section{Challenges and Future Directions} 
\subsection{More Advanced Methods}
The research on model compression techniques for LLMs is still in its early stages. These compressed LLMs, as demonstrated in prior studies \cite{FrantarA23,abs-2305-17888,HoSY23}, continue to exhibit a significant performance gap when compared to their uncompressed counterparts. By delving into more advanced model compression methods tailored for LLMs, we have the potential to enhance the performance of these uncompressed LLMs.
 
\subsection{Scaling up Model Compression Methods from Other Models}
In our paper, we introduce several representative model compression methods for LLMs. However, many classic model compression methods remain prevalent in traditional small models. For example, lottery tickets~\cite{FrankleC19} and parameter sharing~\cite{SavareseM19} are widely used model compression methods in small models. These methods still hold significant potential in the era of LLMs. Future work should focus on exploring how to extend these compression methods to LLMs to achieve further compression.

\subsection{LLM Inference and Deployment}
The efficiency of compressed LLMs during deployment is also a significant area for exploration. This involves multiple evaluation metrics, including arithmetic intensity, memory size, and throughput. Furthermore, we can use an analytical tool, the Roofline Model~\cite{WilliamsWP09}, to assess the resource efficiency of compressed LLMs on specific hardware. Evaluating the deployment efficiency of compressed LLMs on specific hardware can guide researchers in selecting and analyzing the advantages and disadvantages of various model compression methods and further optimizing these methods.

\subsection{The Effect of Scaling Law}
The scaling law~\cite{abs-2001-08361} underscores the significant impact of model size, dataset size, and compute resources on the performance of LLMs. However, the scaling law presents a fundamental challenge for LLM compression, i.e.,  there is a trade-off between model size and performance in compressed LLMs. Delving into the mechanisms and theories underpinning the scaling law is crucial for elucidating and potentially overcoming this limitation.

\subsection{AutoML for LLM Compression}
Existing compression techniques have made remarkable progress, but they still heavily depend on manual design. For instance, designing  appropriate student architectures for knowledge distillation requires a significant amount of human effort. To reduce this reliance on manual design, a feasible solution is to combine Automated Machine Learning (AutoML) techniques such as Meta-Learning~\cite{FinnAL17} and Neural Architecture Search (NAS)~\cite{ZophL17} with model compression. By combining with AutoML techniques, model compression can automatically select appropriate hyperparameters and  tailor  architectures and scales of compressed models,  thus minimizing human involvement and lowering the associated costs. Furthermore, AutoML can identify optimal model compression strategies tailored to specific task requirements, thereby further enhancing compression rates without compromising model performance.

\subsection{Explainability of LLM Compression} 
 Earlier research~\cite{StantonIKAW21,XuZG0MW21} has raised significant concerns regarding the explainability of model compression techniques applied to Pre-trained Language Models (PLMs). Notably, these same challenges extend to LLM compression methods as well. For example, CoT-distillation can enhance SLMs' reasoning performance, yet the mechanism through which it imparts CoT ability remains unclear. This challenge underscores the importance of integrating explainability with model compression approaches for the advancement of LLM compression applications. Explainability not only clarifies the changes and trade-offs in the compression process but also enhances efficiency and accuracy. Additionally, interpretability aids in evaluating the compressed model's performance to ensure it aligns with practical requirements.

% As LLM compression advances, there's a clear call for research into advanced methodologies specifically for LLMs, unlocking their potential across applications.
 
\section{Conclusion}
 In the survey, we have explored model compression techniques for LLMs. Our coverage spanned compression methods, metrics, and benchmark datasets. By diving into LLM compression, we've highlighted its challenges and opportunities.  This survey aims to be a valuable reference, providing insights into the current landscape and promoting ongoing exploration of this pivotal topic.

\section*{Acknowledgments}
We would like to thank the anonymous reviewers and the Action Editor for their valuable feedback and discussions. The work of Jian Li is supported partially by National Natural Science Foundation of China (No.\ 62106257). The work of Yong Liu is supported partially by National Natural Science Foundation of China (No.\ 62076234), Beijing Outstanding Young Scientist Program (No.\ BJJWZYJH012019100020098), the Unicom Innovation Ecological Cooperation Plan, and the CCF-Huawei Populus Grove Fund.

\bibliography{tacl2021}

\begin{thebibliography}{112}
\expandafter\ifx\csname natexlab\endcsname\relax\def\natexlab#1{#1}\fi

\bibitem[{Agarwal et~al.(2024)Agarwal, Vieillard, Zhou, Stanczyk, Garea, Geist, and Bachem}]{agarwal2024generalized}
Rishabh Agarwal, Nino Vieillard, Yongchao Zhou, Piotr Stanczyk, Sabela~Ramos Garea, Matthieu Geist, and Olivier Bachem. 2024.
\newblock \href {https://openreview.net/forum?id=3zKtaqxLhW} {Generalized knowledge distillation for auto-regressive language models}.
\newblock In \emph{The Twelfth International Conference on Learning Representations}.

\bibitem[{An et~al.(2024)An, Zhao, Yu, Tang, and Wang}]{AnZYTW24}
Yongqi An, Xu~Zhao, Tao Yu, Ming Tang, and Jinqiao Wang. 2024.
\newblock \href {https://doi.org/10.1609/AAAI.V38I10.28960} {Fluctuation-based adaptive structured pruning for large language models}.
\newblock In \emph{Thirty-Eighth {AAAI} Conference on Artificial Intelligence, {AAAI} 2024, Thirty-Sixth Conference on Innovative Applications of Artificial Intelligence, {IAAI} 2024, Fourteenth Symposium on Educational Advances in Artificial Intelligence, {EAAI} 2014, February 20-27, 2024, Vancouver, Canada}, pages 10865--10873. {AAAI} Press.

\bibitem[{Ashkboos et~al.(2024)Ashkboos, Croci, do~Nascimento, Hoefler, and Hensman}]{ashkboos2024slicegpt}
Saleh Ashkboos, Maximilian~L. Croci, Marcelo~Gennari do~Nascimento, Torsten Hoefler, and James Hensman. 2024.
\newblock \href {https://openreview.net/forum?id=vXxardq6db} {Slice{GPT}: Compress large language models by deleting rows and columns}.
\newblock In \emph{The Twelfth International Conference on Learning Representations}.

\bibitem[{Brooks et~al.(2023)Brooks, Holynski, and Efros}]{brooks2023instructpix2pix}
Tim Brooks, Aleksander Holynski, and Alexei~A. Efros. 2023.
\newblock \href {https://doi.org/10.1109/CVPR52729.2023.01764} {Instructpix2pix: Learning to follow image editing instructions}.
\newblock In \emph{{IEEE/CVF} Conference on Computer Vision and Pattern Recognition, {CVPR} 2023, Vancouver, BC, Canada, June 17-24, 2023}, pages 18392--18402. {IEEE}.

\bibitem[{Brown et~al.(2020)Brown, Mann, Ryder, Subbiah, Kaplan, Dhariwal, Neelakantan, Shyam, Sastry, Askell, Agarwal, Herbert{-}Voss, Krueger, Henighan, Child, Ramesh, Ziegler, Wu, Winter, Hesse, Chen, Sigler, Litwin, Gray, Chess, Clark, Berner, McCandlish, Radford, Sutskever, and Amodei}]{BrownMRSKDNSSAA20}
Tom~B. Brown, Benjamin Mann, Nick Ryder, Melanie Subbiah, Jared Kaplan, Prafulla Dhariwal, Arvind Neelakantan, Pranav Shyam, Girish Sastry, Amanda Askell, Sandhini Agarwal, Ariel Herbert{-}Voss, Gretchen Krueger, Tom Henighan, Rewon Child, Aditya Ramesh, Daniel~M. Ziegler, Jeffrey Wu, Clemens Winter, Christopher Hesse, Mark Chen, Eric Sigler, Mateusz Litwin, Scott Gray, Benjamin Chess, Jack Clark, Christopher Berner, Sam McCandlish, Alec Radford, Ilya Sutskever, and Dario Amodei. 2020.
\newblock \href {https://proceedings.neurips.cc/paper/2020/hash/1457c0d6bfcb4967418bfb8ac142f64a-Abstract.html} {Language models are few-shot learners}.
\newblock In \emph{Advances in Neural Information Processing Systems 33: Annual Conference on Neural Information Processing Systems 2020, NeurIPS 2020, December 6-12, 2020, virtual}.

\bibitem[{Chee et~al.(2023)Chee, Cai, Kuleshov, and Sa}]{chee2023quip}
Jerry Chee, Yaohui Cai, Volodymyr Kuleshov, and Christopher~De Sa. 2023.
\newblock \href {https://openreview.net/forum?id=xrk9g5vcXR} {Qu{IP}: 2-bit quantization of large language models with guarantees}.
\newblock In \emph{Thirty-seventh Conference on Neural Information Processing Systems}.

\bibitem[{Choquette et~al.(2021)Choquette, Gandhi, Giroux, Stam, and Krashinsky}]{ChoquetteGGSK21}
Jack Choquette, Wishwesh Gandhi, Olivier Giroux, Nick Stam, and Ronny Krashinsky. 2021.
\newblock \href {https://doi.org/10.1109/MM.2021.3061394} {{NVIDIA} {A100} tensor core {GPU:} performance and innovation}.
\newblock \emph{{IEEE} Micro}, 41(2):29--35.

\bibitem[{Chung et~al.(2024)Chung, Hou, Longpre, Zoph, Tay, Fedus, Li, Wang, Dehghani, Brahma, Webson, Gu, Dai, Suzgun, Chen, Chowdhery, Castro-Ros, Pellat, Robinson, Valter, Narang, Mishra, Yu, Zhao, Huang, Dai, Yu, Petrov, Chi, Dean, Devlin, Roberts, Zhou, Le, and Wei}]{abs-2210-11416}
Hyung~Won Chung, Le~Hou, Shayne Longpre, Barret Zoph, Yi~Tay, William Fedus, Yunxuan Li, Xuezhi Wang, Mostafa Dehghani, Siddhartha Brahma, Albert Webson, Shixiang~Shane Gu, Zhuyun Dai, Mirac Suzgun, Xinyun Chen, Aakanksha Chowdhery, Alex Castro-Ros, Marie Pellat, Kevin Robinson, Dasha Valter, Sharan Narang, Gaurav Mishra, Adams Yu, Vincent Zhao, Yanping Huang, Andrew Dai, Hongkun Yu, Slav Petrov, Ed~H. Chi, Jeff Dean, Jacob Devlin, Adam Roberts, Denny Zhou, Quoc~V. Le, and Jason Wei. 2024.
\newblock \href {http://jmlr.org/papers/v25/23-0870.html} {Scaling instruction-finetuned language models}.
\newblock \emph{Journal of Machine Learning Research}, 25(70):1--53.

\bibitem[{Cobbe et~al.(2021)Cobbe, Kosaraju, Bavarian, Chen, Jun, Kaiser, Plappert, Tworek, Hilton, Nakano, Hesse, and Schulman}]{abs-2110-14168}
Karl Cobbe, Vineet Kosaraju, Mohammad Bavarian, Mark Chen, Heewoo Jun, Lukasz Kaiser, Matthias Plappert, Jerry Tworek, Jacob Hilton, Reiichiro Nakano, Christopher Hesse, and John Schulman. 2021.
\newblock \href {http://arxiv.org/abs/2110.14168} {Training verifiers to solve math word problems}.
\newblock \emph{CoRR}, abs/2110.14168.

\bibitem[{Dettmers et~al.(2022)Dettmers, Lewis, Belkada, and Zettlemoyer}]{DettmersLBZ22}
Tim Dettmers, Mike Lewis, Younes Belkada, and Luke Zettlemoyer. 2022.
\newblock \href {http://papers.nips.cc/paper\_files/paper/2022/hash/c3ba4962c05c49636d4c6206a97e9c8a-Abstract-Conference.html} {Gpt3.int8(): 8-bit matrix multiplication for transformers at scale}.
\newblock In \emph{Advances in Neural Information Processing Systems 35: Annual Conference on Neural Information Processing Systems 2022, NeurIPS 2022, New Orleans, LA, USA, November 28 - December 9, 2022}.

\bibitem[{Dettmers et~al.(2023)Dettmers, Pagnoni, Holtzman, and Zettlemoyer}]{DettmersPHZ23}
Tim Dettmers, Artidoro Pagnoni, Ari Holtzman, and Luke Zettlemoyer. 2023.
\newblock \href {http://papers.nips.cc/paper\_files/paper/2023/hash/1feb87871436031bdc0f2beaa62a049b-Abstract-Conference.html} {Qlora: Efficient finetuning of quantized llms}.
\newblock In \emph{Advances in Neural Information Processing Systems 36: Annual Conference on Neural Information Processing Systems 2023, NeurIPS 2023, New Orleans, LA, USA, December 10 - 16, 2023}.

\bibitem[{Dettmers et~al.(2024)Dettmers, Svirschevski, Egiazarian, Kuznedelev, Frantar, Ashkboos, Borzunov, Hoefler, and Alistarh}]{dettmers2024spqr}
Tim Dettmers, Ruslan~A. Svirschevski, Vage Egiazarian, Denis Kuznedelev, Elias Frantar, Saleh Ashkboos, Alexander Borzunov, Torsten Hoefler, and Dan Alistarh. 2024.
\newblock \href {https://openreview.net/forum?id=Q1u25ahSuy} {Sp{QR}: A sparse-quantized representation for near-lossless {LLM} weight compression}.
\newblock In \emph{The Twelfth International Conference on Learning Representations}.

\bibitem[{Dong et~al.(2023)Dong, Li, Dai, Zheng, Wu, Chang, Sun, Xu, Li, and Sui}]{abs-2301-00234}
Qingxiu Dong, Lei Li, Damai Dai, Ce~Zheng, Zhiyong Wu, Baobao Chang, Xu~Sun, Jingjing Xu, Lei Li, and Zhifang Sui. 2023.
\newblock \href {https://doi.org/10.48550/arXiv.2301.00234} {A survey for in-context learning}.
\newblock \emph{CoRR}, abs/2301.00234.

\bibitem[{Du et~al.(2024)Du, Zhang, Cao, Guo, Cao, Chu, and Xu}]{abs-2402-10631}
Dayou Du, Yijia Zhang, Shijie Cao, Jiaqi Guo, Ting Cao, Xiaowen Chu, and Ningyi Xu. 2024.
\newblock \href {https://doi.org/10.48550/ARXIV.2402.10631} {Bitdistiller: Unleashing the potential of sub-4-bit llms via self-distillation}.
\newblock \emph{CoRR}, abs/2402.10631.

\bibitem[{Finn et~al.(2017)Finn, Abbeel, and Levine}]{FinnAL17}
Chelsea Finn, Pieter Abbeel, and Sergey Levine. 2017.
\newblock \href {http://proceedings.mlr.press/v70/finn17a.html} {Model-agnostic meta-learning for fast adaptation of deep networks}.
\newblock In \emph{Proceedings of the 34th International Conference on Machine Learning, {ICML} 2017, Sydney, NSW, Australia, 6-11 August 2017}, volume~70 of \emph{Proceedings of Machine Learning Research}, pages 1126--1135. {PMLR}.

\bibitem[{Frankle and Carbin(2019)}]{FrankleC19}
Jonathan Frankle and Michael Carbin. 2019.
\newblock \href {https://openreview.net/forum?id=rJl-b3RcF7} {The lottery ticket hypothesis: Finding sparse, trainable neural networks}.
\newblock In \emph{7th International Conference on Learning Representations, {ICLR} 2019, New Orleans, LA, USA, May 6-9, 2019}. OpenReview.net.

\bibitem[{Frantar and Alistarh(2022)}]{frantar2022optimal}
Elias Frantar and Dan Alistarh. 2022.
\newblock \href {https://openreview.net/forum?id=ksVGCOlOEba} {Optimal brain compression: A framework for accurate post-training quantization and pruning}.
\newblock In \emph{Advances in Neural Information Processing Systems}.

\bibitem[{Frantar and Alistarh(2023)}]{FrantarA23}
Elias Frantar and Dan Alistarh. 2023.
\newblock \href {https://proceedings.mlr.press/v202/frantar23a.html} {Sparsegpt: Massive language models can be accurately pruned in one-shot}.
\newblock In \emph{International Conference on Machine Learning, {ICML} 2023, 23-29 July 2023, Honolulu, Hawaii, {USA}}, volume 202 of \emph{Proceedings of Machine Learning Research}, pages 10323--10337. {PMLR}.

\bibitem[{Frantar et~al.(2023)Frantar, Ashkboos, Hoefler, and Alistarh}]{frantar2023optq}
Elias Frantar, Saleh Ashkboos, Torsten Hoefler, and Dan Alistarh. 2023.
\newblock \href {https://openreview.net/forum?id=tcbBPnfwxS} {{OPTQ}: Accurate quantization for generative pre-trained transformers}.
\newblock In \emph{The Eleventh International Conference on Learning Representations}.

\bibitem[{Fu et~al.(2023)Fu, Peng, Ou, Sabharwal, and Khot}]{pmlr-v202-fu23d}
Yao Fu, Hao Peng, Litu Ou, Ashish Sabharwal, and Tushar Khot. 2023.
\newblock \href {https://proceedings.mlr.press/v202/fu23d.html} {Specializing smaller language models towards multi-step reasoning}.
\newblock In \emph{Proceedings of the 40th International Conference on Machine Learning}, volume 202 of \emph{Proceedings of Machine Learning Research}, pages 10421--10430. PMLR.

\bibitem[{Gao et~al.(2023)Gao, Tow, Abbasi, Biderman, Black, DiPofi, Foster, Golding, Hsu, Le~Noac'h, Li, McDonell, Muennighoff, Ociepa, Phang, Reynolds, Schoelkopf, Skowron, Sutawika, Tang, Thite, Wang, Wang, and Zou}]{eval-harness}
Leo Gao, Jonathan Tow, Baber Abbasi, Stella Biderman, Sid Black, Anthony DiPofi, Charles Foster, Laurence Golding, Jeffrey Hsu, Alain Le~Noac'h, Haonan Li, Kyle McDonell, Niklas Muennighoff, Chris Ociepa, Jason Phang, Laria Reynolds, Hailey Schoelkopf, Aviya Skowron, Lintang Sutawika, Eric Tang, Anish Thite, Ben Wang, Kevin Wang, and Andy Zou. 2023.
\newblock \href {https://doi.org/10.5281/zenodo.10256836} {A framework for few-shot language model evaluation}.

\bibitem[{Geva et~al.(2021)Geva, Khashabi, Segal, Khot, Roth, and Berant}]{GevaKSKRB21}
Mor Geva, Daniel Khashabi, Elad Segal, Tushar Khot, Dan Roth, and Jonathan Berant. 2021.
\newblock \href {https://doi.org/10.1162/tacl\_a\_00370} {Did aristotle use a laptop? {A} question answering benchmark with implicit reasoning strategies}.
\newblock \emph{Trans. Assoc. Comput. Linguistics}, 9:346--361.

\bibitem[{Gray and Neuhoff(1998)}]{720541}
R.M. Gray and D.L. Neuhoff. 1998.
\newblock \href {https://doi.org/10.1109/18.720541} {Quantization}.
\newblock \emph{IEEE Transactions on Information Theory}, 44(6):2325--2383.

\bibitem[{Gu et~al.(2024)Gu, Dong, Wei, and Huang}]{gu2024minillm}
Yuxian Gu, Li~Dong, Furu Wei, and Minlie Huang. 2024.
\newblock \href {https://openreview.net/forum?id=5h0qf7IBZZ} {Mini{LLM}: Knowledge distillation of large language models}.
\newblock In \emph{The Twelfth International Conference on Learning Representations}.

\bibitem[{Guo et~al.(2023)Guo, Tang, Hu, Leng, Zhang, Yang, Liu, Guo, and Zhu}]{0003THL00LG023}
Cong Guo, Jiaming Tang, Weiming Hu, Jingwen Leng, Chen Zhang, Fan Yang, Yunxin Liu, Minyi Guo, and Yuhao Zhu. 2023.
\newblock \href {https://doi.org/10.1145/3579371.3589038} {Olive: Accelerating large language models via hardware-friendly outlier-victim pair quantization}.
\newblock In \emph{Proceedings of the 50th Annual International Symposium on Computer Architecture, {ISCA} 2023, Orlando, FL, USA, June 17-21, 2023}, pages 3:1--3:15. {ACM}.

\bibitem[{Han et~al.(2016)Han, Mao, and Dally}]{HanMD15}
Song Han, Huizi Mao, and William~J. Dally. 2016.
\newblock \href {http://arxiv.org/abs/1510.00149} {Deep compression: Compressing deep neural network with pruning, trained quantization and huffman coding}.
\newblock In \emph{4th International Conference on Learning Representations, {ICLR} 2016, San Juan, Puerto Rico, May 2-4, 2016, Conference Track Proceedings}.

\bibitem[{Han et~al.(2015)Han, Pool, Tran, and Dally}]{NIPS2015_ae0eb3ee}
Song Han, Jeff Pool, John Tran, and William Dally. 2015.
\newblock \href {https://proceedings.neurips.cc/paper_files/paper/2015/file/ae0eb3eed39d2bcef4622b2499a05fe6-Paper.pdf} {Learning both weights and connections for efficient neural network}.
\newblock In \emph{Advances in Neural Information Processing Systems}, volume~28. Curran Associates, Inc.

\bibitem[{Hinton et~al.(2015)Hinton, Vinyals, and Dean}]{HintonVD15}
Geoffrey~E. Hinton, Oriol Vinyals, and Jeffrey Dean. 2015.
\newblock \href {http://arxiv.org/abs/1503.02531} {Distilling the knowledge in a neural network}.
\newblock \emph{CoRR}, abs/1503.02531.

\bibitem[{Ho et~al.(2023)Ho, Schmid, and Yun}]{HoSY23}
Namgyu Ho, Laura Schmid, and Se{-}Young Yun. 2023.
\newblock \href {https://aclanthology.org/2023.acl-long.830} {Large language models are reasoning teachers}.
\newblock In \emph{Proceedings of the 61st Annual Meeting of the Association for Computational Linguistics (Volume 1: Long Papers), {ACL} 2023, Toronto, Canada, July 9-14, 2023}, pages 14852--14882. Association for Computational Linguistics.

\bibitem[{Hooper et~al.(2024)Hooper, Kim, Mohammadzadeh, Mahoney, Shao, Keutzer, and Gholami}]{abs-2401-18079}
Coleman Hooper, Sehoon Kim, Hiva Mohammadzadeh, Michael~W. Mahoney, Yakun~Sophia Shao, Kurt Keutzer, and Amir Gholami. 2024.
\newblock \href {https://doi.org/10.48550/ARXIV.2401.18079} {Kvquant: Towards 10 million context length {LLM} inference with {KV} cache quantization}.
\newblock \emph{CoRR}, abs/2401.18079.

\bibitem[{Hsieh et~al.(2023)Hsieh, Li, Yeh, Nakhost, Fujii, Ratner, Krishna, Lee, and Pfister}]{HsiehLYNFRKLP23}
Cheng{-}Yu Hsieh, Chun{-}Liang Li, Chih{-}Kuan Yeh, Hootan Nakhost, Yasuhisa Fujii, Alex Ratner, Ranjay Krishna, Chen{-}Yu Lee, and Tomas Pfister. 2023.
\newblock \href {https://aclanthology.org/2023.findings-acl.507} {Distilling step-by-step! outperforming larger language models with less training data and smaller model sizes}.
\newblock In \emph{Findings of the Association for Computational Linguistics: {ACL} 2023, Toronto, Canada, July 9-14, 2023}, pages 8003--8017. Association for Computational Linguistics.

\bibitem[{Huang et~al.(2022)Huang, Chen, Yu, and McKeown}]{abs-2212-10670}
Yukun Huang, Yanda Chen, Zhou Yu, and Kathleen~R. McKeown. 2022.
\newblock \href {https://doi.org/10.48550/arXiv.2212.10670} {In-context learning distillation: Transferring few-shot learning ability of pre-trained language models}.
\newblock \emph{CoRR}, abs/2212.10670.

\bibitem[{Jeon et~al.(2024)Jeon, Kim, and Kim}]{abs-2402-04902}
Hyesung Jeon, Yulhwa Kim, and Jae{-}Joon Kim. 2024.
\newblock \href {https://doi.org/10.48550/ARXIV.2402.04902} {{L4Q:} parameter efficient quantization-aware training on large language models via lora-wise {LSQ}}.
\newblock \emph{CoRR}, abs/2402.04902.

\bibitem[{Jiang et~al.(2023)Jiang, Chan, Chen, and Wang}]{jiang-etal-2023-lion}
Yuxin Jiang, Chunkit Chan, Mingyang Chen, and Wei Wang. 2023.
\newblock \href {https://doi.org/10.18653/v1/2023.emnlp-main.189} {Lion: Adversarial distillation of proprietary large language models}.
\newblock In \emph{Proceedings of the 2023 Conference on Empirical Methods in Natural Language Processing}, pages 3134--3154, Singapore. Association for Computational Linguistics.

\bibitem[{Kaplan et~al.(2020)Kaplan, McCandlish, Henighan, Brown, Chess, Child, Gray, Radford, Wu, and Amodei}]{abs-2001-08361}
Jared Kaplan, Sam McCandlish, Tom Henighan, Tom~B. Brown, Benjamin Chess, Rewon Child, Scott Gray, Alec Radford, Jeffrey Wu, and Dario Amodei. 2020.
\newblock \href {http://arxiv.org/abs/2001.08361} {Scaling laws for neural language models}.
\newblock \emph{CoRR}, abs/2001.08361.

\bibitem[{Kim et~al.(2024)Kim, Kim, Kim, Castells, Choi, Shin, and Song}]{kim2024mefomo}
Bo-Kyeong Kim, Geonmin Kim, Tae-Ho Kim, Thibault Castells, Shinkook Choi, Junho Shin, and Hyoung-Kyu Song. 2024.
\newblock \href {https://openreview.net/forum?id=18VGxuOdpu} {Shortened llama: A simple depth pruning for large language models}.
\newblock \emph{ICLR Workshop on Mathematical and Empirical Understanding of Foundation Models (ME-FoMo)}.

\bibitem[{Kim et~al.(2023{\natexlab{a}})Kim, Lee, Kim, Park, Yoo, Kwon, and Lee}]{abs-2305-14152}
Jeonghoon Kim, Jung~Hyun Lee, Sungdong Kim, Joonsuk Park, Kang~Min Yoo, Se~Jung Kwon, and Dongsoo Lee. 2023{\natexlab{a}}.
\newblock \href {https://openreview.net/forum?id=2jUKhUrBxP} {Memory-efficient fine-tuning of compressed large language models via sub-4-bit integer quantization}.
\newblock In \emph{Thirty-seventh Conference on Neural Information Processing Systems}.

\bibitem[{Kim et~al.(2023{\natexlab{b}})Kim, Hooper, Gholami, Dong, Li, Shen, Mahoney, and Keutzer}]{abs-2306-07629}
Sehoon Kim, Coleman Hooper, Amir Gholami, Zhen Dong, Xiuyu Li, Sheng Shen, Michael~W. Mahoney, and Kurt Keutzer. 2023{\natexlab{b}}.
\newblock \href {https://doi.org/10.48550/ARXIV.2306.07629} {Squeezellm: Dense-and-sparse quantization}.
\newblock \emph{CoRR}, abs/2306.07629.

\bibitem[{LeCun et~al.(1989)LeCun, Denker, and Solla}]{CunDS89}
Yann LeCun, John~S. Denker, and Sara~A. Solla. 1989.
\newblock \href {http://papers.nips.cc/paper/250-optimal-brain-damage} {Optimal brain damage}.
\newblock In \emph{Advances in Neural Information Processing Systems 2, {[NIPS} Conference, Denver, Colorado, USA, November 27-30, 1989]}, pages 598--605. Morgan Kaufmann.

\bibitem[{Lee et~al.(2024)Lee, Jin, Kim, Kim, and Park}]{LeeJKKP24}
Changhun Lee, Jungyu Jin, Taesu Kim, Hyungjun Kim, and Eunhyeok Park. 2024.
\newblock \href {https://doi.org/10.1609/AAAI.V38I12.29237} {{OWQ:} outlier-aware weight quantization for efficient fine-tuning and inference of large language models}.
\newblock In \emph{Thirty-Eighth {AAAI} Conference on Artificial Intelligence, {AAAI} 2024, Thirty-Sixth Conference on Innovative Applications of Artificial Intelligence, {IAAI} 2024, Fourteenth Symposium on Educational Advances in Artificial Intelligence, {EAAI} 2014, February 20-27, 2024, Vancouver, Canada}, pages 13355--13364. {AAAI} Press.

\bibitem[{Li et~al.(2024{\natexlab{a}})Li, Chen, Chen, He, Gu, and Zhou}]{abs-2402-10110}
Ming Li, Lichang Chen, Jiuhai Chen, Shwai He, Jiuxiang Gu, and Tianyi Zhou. 2024{\natexlab{a}}.
\newblock \href {https://doi.org/10.48550/ARXIV.2402.10110} {Selective reflection-tuning: Student-selected data recycling for {LLM} instruction-tuning}.
\newblock \emph{CoRR}, abs/2402.10110.

\bibitem[{Li et~al.(2024{\natexlab{b}})Li, Chen, yelong shen, Chen, Zhang, Li, Wang, Qian, Peng, Mao, Chen, and Yan}]{li2024explanations}
Shiyang Li, Jianshu Chen, yelong shen, Zhiyu Chen, Xinlu Zhang, Zekun Li, Hong Wang, Jing Qian, Baolin Peng, Yi~Mao, Wenhu Chen, and Xifeng Yan. 2024{\natexlab{b}}.
\newblock \href {https://openreview.net/forum?id=rH8ZUcfL9r} {Explanations from large language models make small reasoners better}.
\newblock In \emph{2nd Workshop on Sustainable AI}.

\bibitem[{Li et~al.(2024{\natexlab{c}})Li, Yuan, Feng, Pan, Sun, Wang, Wang, and Li}]{LiYFPSWW024}
Yiwei Li, Peiwen Yuan, Shaoxiong Feng, Boyuan Pan, Bin Sun, Xinglin Wang, Heda Wang, and Kan Li. 2024{\natexlab{c}}.
\newblock \href {https://doi.org/10.1609/AAAI.V38I17.29821} {Turning dust into gold: Distilling complex reasoning capabilities from llms by leveraging negative data}.
\newblock In \emph{Thirty-Eighth {AAAI} Conference on Artificial Intelligence, {AAAI} 2024, Thirty-Sixth Conference on Innovative Applications of Artificial Intelligence, {IAAI} 2024, Fourteenth Symposium on Educational Advances in Artificial Intelligence, {EAAI} 2014, February 20-27, 2024, Vancouver, Canada}, pages 18591--18599. {AAAI} Press.

\bibitem[{Li et~al.(2023{\natexlab{a}})Li, Yu, Liang, He, Karampatziakis, Chen, and Zhao}]{abs-2310-08659}
Yixiao Li, Yifan Yu, Chen Liang, Pengcheng He, Nikos Karampatziakis, Weizhu Chen, and Tuo Zhao. 2023{\natexlab{a}}.
\newblock \href {https://doi.org/10.48550/ARXIV.2310.08659} {Loftq: Lora-fine-tuning-aware quantization for large language models}.
\newblock \emph{CoRR}, abs/2310.08659.

\bibitem[{Li et~al.(2023{\natexlab{b}})Li, Niu, Zhang, Liu, Zhu, and Kang}]{abs-2310-15929}
Yun Li, Lin Niu, Xipeng Zhang, Kai Liu, Jianchen Zhu, and Zhanhui Kang. 2023{\natexlab{b}}.
\newblock \href {https://doi.org/10.48550/ARXIV.2310.15929} {E-sparse: Boosting the large language model inference through entropy-based {N:} {M} sparsity}.
\newblock \emph{CoRR}, abs/2310.15929.

\bibitem[{Li et~al.(2023{\natexlab{c}})Li, Li, and Meng}]{LiLM23}
Zhuo Li, Hengyi Li, and Lin Meng. 2023{\natexlab{c}}.
\newblock \href {https://doi.org/10.3390/COMPUTERS12030060} {Model compression for deep neural networks: {A} survey}.
\newblock \emph{Comput.}, 12(3):60.

\bibitem[{Liang et~al.(2023)Liang, Zuo, Zhang, He, Chen, and Zhao}]{LiangZZHCZ23}
Chen Liang, Simiao Zuo, Qingru Zhang, Pengcheng He, Weizhu Chen, and Tuo Zhao. 2023.
\newblock \href {https://proceedings.mlr.press/v202/liang23j.html} {Less is more: Task-aware layer-wise distillation for language model compression}.
\newblock In \emph{International Conference on Machine Learning, {ICML} 2023, 23-29 July 2023, Honolulu, Hawaii, {USA}}, volume 202 of \emph{Proceedings of Machine Learning Research}, pages 20852--20867. {PMLR}.

\bibitem[{Lin et~al.(2023)Lin, Tang, Tang, Yang, Dang, and Han}]{abs-2306-00978}
Ji~Lin, Jiaming Tang, Haotian Tang, Shang Yang, Xingyu Dang, and Song Han. 2023.
\newblock \href {https://doi.org/10.48550/arXiv.2306.00978} {{AWQ:} activation-aware weight quantization for {LLM} compression and acceleration}.
\newblock \emph{CoRR}, abs/2306.00978.

\bibitem[{Liu et~al.(2023{\natexlab{a}})Liu, Liu, Huang, Dong, and Cheng}]{liu-etal-2023-llm}
Shih-yang Liu, Zechun Liu, Xijie Huang, Pingcheng Dong, and Kwang-Ting Cheng. 2023{\natexlab{a}}.
\newblock \href {https://doi.org/10.18653/v1/2023.emnlp-main.39} {{LLM}-{FP}4: 4-bit floating-point quantized transformers}.
\newblock In \emph{Proceedings of the 2023 Conference on Empirical Methods in Natural Language Processing}, pages 592--605, Singapore. Association for Computational Linguistics.

\bibitem[{Liu(2024)}]{liu2024learning}
Yuxuan Liu. 2024.
\newblock \href {https://openreview.net/forum?id=auvDeqEKrk} {Learning to reason with autoregressive in-context distillation}.
\newblock In \emph{The Second Tiny Papers Track at ICLR 2024}.

\bibitem[{Liu et~al.(2023{\natexlab{b}})Liu, Oguz, Zhao, Chang, Stock, Mehdad, Shi, Krishnamoorthi, and Chandra}]{abs-2305-17888}
Zechun Liu, Barlas Oguz, Changsheng Zhao, Ernie Chang, Pierre Stock, Yashar Mehdad, Yangyang Shi, Raghuraman Krishnamoorthi, and Vikas Chandra. 2023{\natexlab{b}}.
\newblock \href {https://doi.org/10.48550/ARXIV.2305.17888} {{LLM-QAT:} data-free quantization aware training for large language models}.
\newblock \emph{CoRR}, abs/2305.17888.

\bibitem[{Liu et~al.(2024)Liu, Yuan, Jin, Zhong, Xu, Braverman, Chen, and Hu}]{abs-2402-02750}
Zirui Liu, Jiayi Yuan, Hongye Jin, Shaochen Zhong, Zhaozhuo Xu, Vladimir Braverman, Beidi Chen, and Xia Hu. 2024.
\newblock \href {https://doi.org/10.48550/ARXIV.2402.02750} {{KIVI:} {A} tuning-free asymmetric 2bit quantization for {KV} cache}.
\newblock \emph{CoRR}, abs/2402.02750.

\bibitem[{Ma et~al.(2023)Ma, Fang, and Wang}]{ma2023llmpruner}
Xinyin Ma, Gongfan Fang, and Xinchao Wang. 2023.
\newblock \href {https://openreview.net/forum?id=J8Ajf9WfXP} {{LLM}-pruner: On the structural pruning of large language models}.
\newblock In \emph{Thirty-seventh Conference on Neural Information Processing Systems}.

\bibitem[{Magister et~al.(2023)Magister, Mallinson, Ad{\'{a}}mek, Malmi, and Severyn}]{MagisterMAMS23}
Lucie~Charlotte Magister, Jonathan Mallinson, Jakub Ad{\'{a}}mek, Eric Malmi, and Aliaksei Severyn. 2023.
\newblock \href {https://aclanthology.org/2023.acl-short.151} {Teaching small language models to reason}.
\newblock In \emph{Proceedings of the 61st Annual Meeting of the Association for Computational Linguistics (Volume 2: Short Papers), {ACL} 2023, Toronto, Canada, July 9-14, 2023}, pages 1773--1781. Association for Computational Linguistics.

\bibitem[{Marcus et~al.(1993)Marcus, Santorini, and Marcinkiewicz}]{10.5555/972470.972475}
Mitchell~P. Marcus, Beatrice Santorini, and Mary~Ann Marcinkiewicz. 1993.
\newblock \href {https://aclanthology.org/J93-2004} {Building a large annotated corpus of {E}nglish: The {P}enn {T}reebank}.
\newblock \emph{Computational Linguistics}, 19(2):313--330.

\bibitem[{Merity et~al.(2017)Merity, Xiong, Bradbury, and Socher}]{MerityX0S17}
Stephen Merity, Caiming Xiong, James Bradbury, and Richard Socher. 2017.
\newblock \href {https://openreview.net/forum?id=Byj72udxe} {Pointer sentinel mixture models}.
\newblock In \emph{5th International Conference on Learning Representations, {ICLR} 2017, Toulon, France, April 24-26, 2017, Conference Track Proceedings}. OpenReview.net.

\bibitem[{Mihaylov et~al.(2018)Mihaylov, Clark, Khot, and Sabharwal}]{MihaylovCKS18}
Todor Mihaylov, Peter Clark, Tushar Khot, and Ashish Sabharwal. 2018.
\newblock \href {https://doi.org/10.18653/V1/D18-1260} {Can a suit of armor conduct electricity? {A} new dataset for open book question answering}.
\newblock In \emph{Proceedings of the 2018 Conference on Empirical Methods in Natural Language Processing, Brussels, Belgium, October 31 - November 4, 2018}, pages 2381--2391. Association for Computational Linguistics.

\bibitem[{Molchanov et~al.(2019)Molchanov, Mallya, Tyree, Frosio, and Kautz}]{MolchanovMTFK19}
Pavlo Molchanov, Arun Mallya, Stephen Tyree, Iuri Frosio, and Jan Kautz. 2019.
\newblock \href {https://doi.org/10.1109/CVPR.2019.01152} {Importance estimation for neural network pruning}.
\newblock In \emph{{IEEE} Conference on Computer Vision and Pattern Recognition, {CVPR} 2019, Long Beach, CA, USA, June 16-20, 2019}, pages 11264--11272. Computer Vision Foundation / {IEEE}.

\bibitem[{OpenAI(2024)}]{openai2024gpt4}
OpenAI. 2024.
\newblock \href {http://arxiv.org/abs/2303.08774} {Gpt-4 technical report}.

\bibitem[{Ouyang et~al.(2022)Ouyang, Wu, Jiang, Almeida, Wainwright, Mishkin, Zhang, Agarwal, Slama, Ray, Schulman, Hilton, Kelton, Miller, Simens, Askell, Welinder, Christiano, Leike, and Lowe}]{Ouyang0JAWMZASR22}
Long Ouyang, Jeffrey Wu, Xu~Jiang, Diogo Almeida, Carroll~L. Wainwright, Pamela Mishkin, Chong Zhang, Sandhini Agarwal, Katarina Slama, Alex Ray, John Schulman, Jacob Hilton, Fraser Kelton, Luke Miller, Maddie Simens, Amanda Askell, Peter Welinder, Paul~F. Christiano, Jan Leike, and Ryan Lowe. 2022.
\newblock \href {http://papers.nips.cc/paper\_files/paper/2022/hash/b1efde53be364a73914f58805a001731-Abstract-Conference.html} {Training language models to follow instructions with human feedback}.
\newblock In \emph{NeurIPS}.

\bibitem[{Paperno et~al.(2016)Paperno, Kruszewski, Lazaridou, Pham, Bernardi, Pezzelle, Baroni, Boleda, and Fern{\'{a}}ndez}]{PapernoKLPBPBBF16}
Denis Paperno, Germ{\'{a}}n Kruszewski, Angeliki Lazaridou, Quan~Ngoc Pham, Raffaella Bernardi, Sandro Pezzelle, Marco Baroni, Gemma Boleda, and Raquel Fern{\'{a}}ndez. 2016.
\newblock \href {https://doi.org/10.18653/v1/p16-1144} {The {LAMBADA} dataset: Word prediction requiring a broad discourse context}.
\newblock In \emph{Proceedings of the 54th Annual Meeting of the Association for Computational Linguistics, {ACL} 2016, August 7-12, 2016, Berlin, Germany, Volume 1: Long Papers}. The Association for Computer Linguistics.

\bibitem[{Park et~al.(2024)Park, park, Kim, Lee, Kim, Kwon, Kwon, Kim, Lee, and Lee}]{park2024lutgemm}
Gunho Park, Baeseong park, Minsub Kim, Sungjae Lee, Jeonghoon Kim, Beomseok Kwon, Se~Jung Kwon, Byeongwook Kim, Youngjoo Lee, and Dongsoo Lee. 2024.
\newblock \href {https://openreview.net/forum?id=gLARhFLE0F} {{LUT}-{GEMM}: Quantized matrix multiplication based on {LUT}s for efficient inference in large-scale generative language models}.
\newblock In \emph{The Twelfth International Conference on Learning Representations}.

\bibitem[{Radford et~al.(2019)Radford, Wu, Child, Luan, Amodei, and Sutskever}]{radford2019language}
Alec Radford, Jeff Wu, Rewon Child, David Luan, Dario Amodei, and Ilya Sutskever. 2019.
\newblock \href {https://cdn.openai.com/better-language-models/language_models_are_unsupervised_multitask_learners.pdf} {Language models are unsupervised multitask learners}.
\newblock \emph{OpenAI blog}, 1(8):9.

\bibitem[{Raffel et~al.(2020)Raffel, Shazeer, Roberts, Lee, Narang, Matena, Zhou, Li, and Liu}]{RaffelSRLNMZLL20}
Colin Raffel, Noam Shazeer, Adam Roberts, Katherine Lee, Sharan Narang, Michael Matena, Yanqi Zhou, Wei Li, and Peter~J. Liu. 2020.
\newblock \href {http://jmlr.org/papers/v21/20-074.html} {Exploring the limits of transfer learning with a unified text-to-text transformer}.
\newblock \emph{J. Mach. Learn. Res.}, 21:140:1--140:67.

\bibitem[{Rastegari et~al.(2016)Rastegari, Ordonez, Redmon, and Farhadi}]{RastegariORF16}
Mohammad Rastegari, Vicente Ordonez, Joseph Redmon, and Ali Farhadi. 2016.
\newblock \href {https://doi.org/10.1007/978-3-319-46493-0\_32} {Xnor-net: Imagenet classification using binary convolutional neural networks}.
\newblock In \emph{Computer Vision - {ECCV} 2016 - 14th European Conference, Amsterdam, The Netherlands, October 11-14, 2016, Proceedings, Part {IV}}, volume 9908 of \emph{Lecture Notes in Computer Science}, pages 525--542. Springer.

\bibitem[{Rogers et~al.(2020)Rogers, Kovaleva, and Rumshisky}]{rogers-etal-2020-primer}
Anna Rogers, Olga Kovaleva, and Anna Rumshisky. 2020.
\newblock \href {https://doi.org/10.1162/tacl_a_00349} {A primer in {BERT}ology: What we know about how {BERT} works}.
\newblock \emph{Transactions of the Association for Computational Linguistics}, 8:842--866.

\bibitem[{Saha et~al.(2023)Saha, Srivastava, and Pilanci}]{SahaSP23}
Rajarshi Saha, Varun Srivastava, and Mert Pilanci. 2023.
\newblock \href {http://papers.nips.cc/paper\_files/paper/2023/hash/3bf4b55960aaa23553cd2a6bdc6e1b57-Abstract-Conference.html} {Matrix compression via randomized low rank and low precision factorization}.
\newblock In \emph{Advances in Neural Information Processing Systems 36: Annual Conference on Neural Information Processing Systems 2023, NeurIPS 2023, New Orleans, LA, USA, December 10 - 16, 2023}.

\bibitem[{Savarese and Maire(2019)}]{SavareseM19}
Pedro Savarese and Michael Maire. 2019.
\newblock \href {https://openreview.net/forum?id=rJgYxn09Fm} {Learning implicitly recurrent cnns through parameter sharing}.
\newblock In \emph{7th International Conference on Learning Representations, {ICLR} 2019, New Orleans, LA, USA, May 6-9, 2019}. OpenReview.net.

\bibitem[{Scao et~al.(2022)Scao, Fan, Akiki, Pavlick, Ilic, Hesslow, Castagn{\'{e}}, Luccioni, Yvon, Gall{\'{e}}, Tow, Rush, Biderman, Webson, Ammanamanchi, Wang, Sagot, Muennighoff, del Moral, Ruwase, Bawden, Bekman, McMillan{-}Major, Beltagy, Nguyen, Saulnier, Tan, Suarez, Sanh, Lauren{\c{c}}on, Jernite, Launay, Mitchell, Raffel, Gokaslan, Simhi, Soroa, Aji, Alfassy, Rogers, Nitzav, Xu, Mou, Emezue, Klamm, Leong, van Strien, Adelani, and et~al.}]{abs-2211-05100}
Teven~Le Scao, Angela Fan, Christopher Akiki, Ellie Pavlick, Suzana Ilic, Daniel Hesslow, Roman Castagn{\'{e}}, Alexandra~Sasha Luccioni, Fran{\c{c}}ois Yvon, Matthias Gall{\'{e}}, Jonathan Tow, Alexander~M. Rush, Stella Biderman, Albert Webson, Pawan~Sasanka Ammanamanchi, Thomas Wang, Beno{\^{\i}}t Sagot, Niklas Muennighoff, Albert~Villanova del Moral, Olatunji Ruwase, Rachel Bawden, Stas Bekman, Angelina McMillan{-}Major, Iz~Beltagy, Huu Nguyen, Lucile Saulnier, Samson Tan, Pedro~Ortiz Suarez, Victor Sanh, Hugo Lauren{\c{c}}on, Yacine Jernite, Julien Launay, Margaret Mitchell, Colin Raffel, Aaron Gokaslan, Adi Simhi, Aitor Soroa, Alham~Fikri Aji, Amit Alfassy, Anna Rogers, Ariel~Kreisberg Nitzav, Canwen Xu, Chenghao Mou, Chris Emezue, Christopher Klamm, Colin Leong, Daniel van Strien, David~Ifeoluwa Adelani, and et~al. 2022.
\newblock \href {https://doi.org/10.48550/ARXIV.2211.05100} {{BLOOM:} {A} 176b-parameter open-access multilingual language model}.
\newblock \emph{CoRR}, abs/2211.05100.

\bibitem[{Shao et~al.(2024{\natexlab{a}})Shao, Liu, and Qian}]{10445737}
Hang Shao, Bei Liu, and Yanmin Qian. 2024{\natexlab{a}}.
\newblock \href {https://doi.org/10.1109/ICASSP48485.2024.10445737} {One-shot sensitivity-aware mixed sparsity pruning for large language models}.
\newblock In \emph{ICASSP 2024 - 2024 IEEE International Conference on Acoustics, Speech and Signal Processing (ICASSP)}, pages 11296--11300.

\bibitem[{Shao et~al.(2024{\natexlab{b}})Shao, Chen, Zhang, Xu, Zhao, Li, Zhang, Gao, Qiao, and Luo}]{shao2024omniquant}
Wenqi Shao, Mengzhao Chen, Zhaoyang Zhang, Peng Xu, Lirui Zhao, Zhiqian Li, Kaipeng Zhang, Peng Gao, Yu~Qiao, and Ping Luo. 2024{\natexlab{b}}.
\newblock \href {https://openreview.net/forum?id=8Wuvhh0LYW} {Omniquant: Omnidirectionally calibrated quantization for large language models}.
\newblock In \emph{The Twelfth International Conference on Learning Representations}.

\bibitem[{Sharma et~al.(2024)Sharma, Ash, and Misra}]{sharma2024the}
Pratyusha Sharma, Jordan~T. Ash, and Dipendra Misra. 2024.
\newblock \href {https://openreview.net/forum?id=ozX92bu8VA} {The truth is in there: Improving reasoning with layer-selective rank reduction}.
\newblock In \emph{The Twelfth International Conference on Learning Representations}.

\bibitem[{Shridhar et~al.(2023)Shridhar, Stolfo, and Sachan}]{ShridharSS23}
Kumar Shridhar, Alessandro Stolfo, and Mrinmaya Sachan. 2023.
\newblock \href {https://aclanthology.org/2023.findings-acl.441} {Distilling reasoning capabilities into smaller language models}.
\newblock In \emph{Findings of the Association for Computational Linguistics: {ACL} 2023, Toronto, Canada, July 9-14, 2023}, pages 7059--7073. Association for Computational Linguistics.

\bibitem[{Srebro and Jaakkola(2003)}]{SrebroJ03}
Nathan Srebro and Tommi~S. Jaakkola. 2003.
\newblock \href {http://www.aaai.org/Library/ICML/2003/icml03-094.php} {Weighted low-rank approximations}.
\newblock In \emph{Machine Learning, Proceedings of the Twentieth International Conference {(ICML} 2003), August 21-24, 2003, Washington, DC, {USA}}, pages 720--727. {AAAI} Press.

\bibitem[{Srivastava et~al.(2023)Srivastava, Rastogi, Rao, Shoeb, Abid, Fisch, Brown, Santoro, Gupta, Garriga-Alonso, Kluska, Lewkowycz, Agarwal, Power, Ray, Warstadt, Kocurek, Safaya, Tazarv, Xiang, Parrish, Nie, Hussain, Askell, Dsouza, Slone, Rahane, Iyer, Andreassen, Madotto, Santilli, Stuhlm{\"u}ller, Dai, La, Lampinen, Zou, Jiang, Chen, Vuong, Gupta, Gottardi, Norelli, Venkatesh, Gholamidavoodi, Tabassum, Menezes, Kirubarajan, Mullokandov, Sabharwal, Herrick, Efrat, Erdem, Karaka{\c{s}}, Roberts, Loe, Zoph, Bojanowski, {\"O}zyurt, Hedayatnia, Neyshabur, Inden, Stein, Ekmekci, Lin, Howald, Orinion, Diao, Dour, Stinson, Argueta, Ferri, Singh, Rathkopf, Meng, Baral, Wu, Callison-Burch, Waites, Voigt, Manning, Potts, Ramirez, Rivera, Siro, Raffel, Ashcraft, Garbacea, Sileo, Garrette, Hendrycks, Kilman, Roth, Freeman, Khashabi, Levy, Gonz{\'a}lez, Perszyk, Hernandez, Chen, Ippolito, Gilboa, Dohan, Drakard, Jurgens, Datta, Ganguli, Emelin, Kleyko, Yuret, Chen, Tam, Hupkes, Misra, Buzan, Mollo, Yang, Lee,
  Schrader, Shutova, Cubuk, Segal, Hagerman, Barnes, Donoway, Pavlick, Rodol{\`a}, Lam, Chu, Tang, Erdem, Chang, Chi, Dyer, Jerzak, Kim, Manyasi, Zheltonozhskii, Xia, Siar, Mart{\'\i}nez-Plumed, Happ{\'e}, Chollet, Rong, Mishra, Winata, de~Melo, Kruszewski, Parascandolo, Mariani, Wang, Jaimovitch-Lopez, Betz, Gur-Ari, Galijasevic, Kim, Rashkin, Hajishirzi, Mehta, Bogar, Shevlin, Schuetze, Yakura, Zhang, Wong, Ng, Noble, Jumelet, Geissinger, Kernion, Hilton, Lee, Fisac, Simon, Koppel, Zheng, Zou, Kocon, Thompson, Wingfield, Kaplan, Radom, Sohl-Dickstein, Phang, Wei, Yosinski, Novikova, Bosscher, Marsh, Kim, Taal, Engel, Alabi, Xu, Song, Tang, Waweru, Burden, Miller, Balis, Batchelder, Berant, Frohberg, Rozen, Hernandez-Orallo, Boudeman, Guerr, Jones, Tenenbaum, Rule, Chua, Kanclerz, Livescu, Krauth, Gopalakrishnan, Ignatyeva, Markert, Dhole, Gimpel, Omondi, Mathewson, Chiafullo, Shkaruta, Shridhar, McDonell, Richardson, Reynolds, Gao, Zhang, Dugan, Qin, Contreras-Ochando, Morency, Moschella, Lam, Noble,
  Schmidt, He, Oliveros-Col{\'o}n, Metz, Senel, Bosma, Sap, Hoeve, Farooqi, Faruqui, Mazeika, Baturan, Marelli, Maru, Ramirez-Quintana, Tolkiehn, Giulianelli, Lewis, Potthast, Leavitt, Hagen, Schubert, Baitemirova, Arnaud, McElrath, Yee, Cohen, Gu, Ivanitskiy, Starritt, Strube, Sw{\k{e}}drowski, Bevilacqua, Yasunaga, Kale, Cain, Xu, Suzgun, Walker, Tiwari, Bansal, Aminnaseri, Geva, Gheini, T, Peng, Chi, Lee, Krakover, Cameron, Roberts, Doiron, Martinez, Nangia, Deckers, Muennighoff, Keskar, Iyer, Constant, Fiedel, Wen, Zhang, Agha, Elbaghdadi, Levy, Evans, Casares, Doshi, Fung, Liang, Vicol, Alipoormolabashi, Liao, Liang, Chang, Eckersley, Htut, Hwang, Mi{\l}kowski, Patil, Pezeshkpour, Oli, Mei, Lyu, Chen, Banjade, Rudolph, Gabriel, Habacker, Risco, Milli{\`e}re, Garg, Barnes, Saurous, Arakawa, Raymaekers, Frank, Sikand, Novak, Sitelew, Bras, Liu, Jacobs, Zhang, Salakhutdinov, Chi, Lee, Stovall, Teehan, Yang, Singh, Mohammad, Anand, Dillavou, Shleifer, Wiseman, Gruetter, Bowman, Schoenholz, Han, Kwatra, Rous,
  Ghazarian, Ghosh, Casey, Bischoff, Gehrmann, Schuster, Sadeghi, Hamdan, Zhou, Srivastava, Shi, Singh, Asaadi, Gu, Pachchigar, Toshniwal, Upadhyay, Debnath, Shakeri, Thormeyer, Melzi, Reddy, Makini, Lee, Torene, Hatwar, Dehaene, Divic, Ermon, Biderman, Lin, Prasad, Piantadosi, Shieber, Misherghi, Kiritchenko, Mishra, Linzen, Schuster, Li, Yu, Ali, Hashimoto, Wu, Desbordes, Rothschild, Phan, Wang, Nkinyili, Schick, Kornev, Tunduny, Gerstenberg, Chang, Neeraj, Khot, Shultz, Shaham, Misra, Demberg, Nyamai, Raunak, Ramasesh, vinay~uday prabhu, Padmakumar, Srikumar, Fedus, Saunders, Zhang, Vossen, Ren, Tong, Zhao, Wu, Shen, Yaghoobzadeh, Lakretz, Song, Bahri, Choi, Yang, Hao, Chen, Belinkov, Hou, Hou, Bai, Seid, Zhao, Wang, Wang, Wang, and Wu}]{srivastava2023beyond}
Aarohi Srivastava, Abhinav Rastogi, Abhishek Rao, Abu Awal~Md Shoeb, Abubakar Abid, Adam Fisch, Adam~R. Brown, Adam Santoro, Aditya Gupta, Adri{\`a} Garriga-Alonso, Agnieszka Kluska, Aitor Lewkowycz, Akshat Agarwal, Alethea Power, Alex Ray, Alex Warstadt, Alexander~W. Kocurek, Ali Safaya, Ali Tazarv, Alice Xiang, Alicia Parrish, Allen Nie, Aman Hussain, Amanda Askell, Amanda Dsouza, Ambrose Slone, Ameet Rahane, Anantharaman~S. Iyer, Anders~Johan Andreassen, Andrea Madotto, Andrea Santilli, Andreas Stuhlm{\"u}ller, Andrew~M. Dai, Andrew La, Andrew Lampinen, Andy Zou, Angela Jiang, Angelica Chen, Anh Vuong, Animesh Gupta, Anna Gottardi, Antonio Norelli, Anu Venkatesh, Arash Gholamidavoodi, Arfa Tabassum, Arul Menezes, Arun Kirubarajan, Asher Mullokandov, Ashish Sabharwal, Austin Herrick, Avia Efrat, Aykut Erdem, Ayla Karaka{\c{s}}, B.~Ryan Roberts, Bao~Sheng Loe, Barret Zoph, Bart{\l}omiej Bojanowski, Batuhan {\"O}zyurt, Behnam Hedayatnia, Behnam Neyshabur, Benjamin Inden, Benno Stein, Berk Ekmekci, Bill~Yuchen
  Lin, Blake Howald, Bryan Orinion, Cameron Diao, Cameron Dour, Catherine Stinson, Cedrick Argueta, Cesar Ferri, Chandan Singh, Charles Rathkopf, Chenlin Meng, Chitta Baral, Chiyu Wu, Chris Callison-Burch, Christopher Waites, Christian Voigt, Christopher~D Manning, Christopher Potts, Cindy Ramirez, Clara~E. Rivera, Clemencia Siro, Colin Raffel, Courtney Ashcraft, Cristina Garbacea, Damien Sileo, Dan Garrette, Dan Hendrycks, Dan Kilman, Dan Roth, C.~Daniel Freeman, Daniel Khashabi, Daniel Levy, Daniel~Mosegu{\'\i} Gonz{\'a}lez, Danielle Perszyk, Danny Hernandez, Danqi Chen, Daphne Ippolito, Dar Gilboa, David Dohan, David Drakard, David Jurgens, Debajyoti Datta, Deep Ganguli, Denis Emelin, Denis Kleyko, Deniz Yuret, Derek Chen, Derek Tam, Dieuwke Hupkes, Diganta Misra, Dilyar Buzan, Dimitri~Coelho Mollo, Diyi Yang, Dong-Ho Lee, Dylan Schrader, Ekaterina Shutova, Ekin~Dogus Cubuk, Elad Segal, Eleanor Hagerman, Elizabeth Barnes, Elizabeth Donoway, Ellie Pavlick, Emanuele Rodol{\`a}, Emma Lam, Eric Chu, Eric Tang,
  Erkut Erdem, Ernie Chang, Ethan~A Chi, Ethan Dyer, Ethan Jerzak, Ethan Kim, Eunice~Engefu Manyasi, Evgenii Zheltonozhskii, Fanyue Xia, Fatemeh Siar, Fernando Mart{\'\i}nez-Plumed, Francesca Happ{\'e}, Francois Chollet, Frieda Rong, Gaurav Mishra, Genta~Indra Winata, Gerard de~Melo, Germ{\'a}n Kruszewski, Giambattista Parascandolo, Giorgio Mariani, Gloria~Xinyue Wang, Gonzalo Jaimovitch-Lopez, Gregor Betz, Guy Gur-Ari, Hana Galijasevic, Hannah Kim, Hannah Rashkin, Hannaneh Hajishirzi, Harsh Mehta, Hayden Bogar, Henry Francis~Anthony Shevlin, Hinrich Schuetze, Hiromu Yakura, Hongming Zhang, Hugh~Mee Wong, Ian Ng, Isaac Noble, Jaap Jumelet, Jack Geissinger, Jackson Kernion, Jacob Hilton, Jaehoon Lee, Jaime~Fern{\'a}ndez Fisac, James~B Simon, James Koppel, James Zheng, James Zou, Jan Kocon, Jana Thompson, Janelle Wingfield, Jared Kaplan, Jarema Radom, Jascha Sohl-Dickstein, Jason Phang, Jason Wei, Jason Yosinski, Jekaterina Novikova, Jelle Bosscher, Jennifer Marsh, Jeremy Kim, Jeroen Taal, Jesse Engel, Jesujoba
  Alabi, Jiacheng Xu, Jiaming Song, Jillian Tang, Joan Waweru, John Burden, John Miller, John~U. Balis, Jonathan Batchelder, Jonathan Berant, J{\"o}rg Frohberg, Jos Rozen, Jose Hernandez-Orallo, Joseph Boudeman, Joseph Guerr, Joseph Jones, Joshua~B. Tenenbaum, Joshua~S. Rule, Joyce Chua, Kamil Kanclerz, Karen Livescu, Karl Krauth, Karthik Gopalakrishnan, Katerina Ignatyeva, Katja Markert, Kaustubh Dhole, Kevin Gimpel, Kevin Omondi, Kory~Wallace Mathewson, Kristen Chiafullo, Ksenia Shkaruta, Kumar Shridhar, Kyle McDonell, Kyle Richardson, Laria Reynolds, Leo Gao, Li~Zhang, Liam Dugan, Lianhui Qin, Lidia Contreras-Ochando, Louis-Philippe Morency, Luca Moschella, Lucas Lam, Lucy Noble, Ludwig Schmidt, Luheng He, Luis Oliveros-Col{\'o}n, Luke Metz, L{\"u}tfi~Kerem Senel, Maarten Bosma, Maarten Sap, Maartje~Ter Hoeve, Maheen Farooqi, Manaal Faruqui, Mantas Mazeika, Marco Baturan, Marco Marelli, Marco Maru, Maria~Jose Ramirez-Quintana, Marie Tolkiehn, Mario Giulianelli, Martha Lewis, Martin Potthast, Matthew~L
  Leavitt, Matthias Hagen, M{\'a}ty{\'a}s Schubert, Medina~Orduna Baitemirova, Melody Arnaud, Melvin McElrath, Michael~Andrew Yee, Michael Cohen, Michael Gu, Michael Ivanitskiy, Michael Starritt, Michael Strube, Micha{\l} Sw{\k{e}}drowski, Michele Bevilacqua, Michihiro Yasunaga, Mihir Kale, Mike Cain, Mimee Xu, Mirac Suzgun, Mitch Walker, Mo~Tiwari, Mohit Bansal, Moin Aminnaseri, Mor Geva, Mozhdeh Gheini, Mukund~Varma T, Nanyun Peng, Nathan~Andrew Chi, Nayeon Lee, Neta Gur-Ari Krakover, Nicholas Cameron, Nicholas Roberts, Nick Doiron, Nicole Martinez, Nikita Nangia, Niklas Deckers, Niklas Muennighoff, Nitish~Shirish Keskar, Niveditha~S. Iyer, Noah Constant, Noah Fiedel, Nuan Wen, Oliver Zhang, Omar Agha, Omar Elbaghdadi, Omer Levy, Owain Evans, Pablo Antonio~Moreno Casares, Parth Doshi, Pascale Fung, Paul~Pu Liang, Paul Vicol, Pegah Alipoormolabashi, Peiyuan Liao, Percy Liang, Peter~W Chang, Peter Eckersley, Phu~Mon Htut, Pinyu Hwang, Piotr Mi{\l}kowski, Piyush Patil, Pouya Pezeshkpour, Priti Oli, Qiaozhu
  Mei, Qing Lyu, Qinlang Chen, Rabin Banjade, Rachel~Etta Rudolph, Raefer Gabriel, Rahel Habacker, Ramon Risco, Rapha{\"e}l Milli{\`e}re, Rhythm Garg, Richard Barnes, Rif~A. Saurous, Riku Arakawa, Robbe Raymaekers, Robert Frank, Rohan Sikand, Roman Novak, Roman Sitelew, Ronan~Le Bras, Rosanne Liu, Rowan Jacobs, Rui Zhang, Russ Salakhutdinov, Ryan~Andrew Chi, Seungjae~Ryan Lee, Ryan Stovall, Ryan Teehan, Rylan Yang, Sahib Singh, Saif~M. Mohammad, Sajant Anand, Sam Dillavou, Sam Shleifer, Sam Wiseman, Samuel Gruetter, Samuel~R. Bowman, Samuel~Stern Schoenholz, Sanghyun Han, Sanjeev Kwatra, Sarah~A. Rous, Sarik Ghazarian, Sayan Ghosh, Sean Casey, Sebastian Bischoff, Sebastian Gehrmann, Sebastian Schuster, Sepideh Sadeghi, Shadi Hamdan, Sharon Zhou, Shashank Srivastava, Sherry Shi, Shikhar Singh, Shima Asaadi, Shixiang~Shane Gu, Shubh Pachchigar, Shubham Toshniwal, Shyam Upadhyay, Shyamolima~Shammie Debnath, Siamak Shakeri, Simon Thormeyer, Simone Melzi, Siva Reddy, Sneha~Priscilla Makini, Soo-Hwan Lee, Spencer
  Torene, Sriharsha Hatwar, Stanislas Dehaene, Stefan Divic, Stefano Ermon, Stella Biderman, Stephanie Lin, Stephen Prasad, Steven Piantadosi, Stuart Shieber, Summer Misherghi, Svetlana Kiritchenko, Swaroop Mishra, Tal Linzen, Tal Schuster, Tao Li, Tao Yu, Tariq Ali, Tatsunori Hashimoto, Te-Lin Wu, Th{\'e}o Desbordes, Theodore Rothschild, Thomas Phan, Tianle Wang, Tiberius Nkinyili, Timo Schick, Timofei Kornev, Titus Tunduny, Tobias Gerstenberg, Trenton Chang, Trishala Neeraj, Tushar Khot, Tyler Shultz, Uri Shaham, Vedant Misra, Vera Demberg, Victoria Nyamai, Vikas Raunak, Vinay~Venkatesh Ramasesh, vinay~uday prabhu, Vishakh Padmakumar, Vivek Srikumar, William Fedus, William Saunders, William Zhang, Wout Vossen, Xiang Ren, Xiaoyu Tong, Xinran Zhao, Xinyi Wu, Xudong Shen, Yadollah Yaghoobzadeh, Yair Lakretz, Yangqiu Song, Yasaman Bahri, Yejin Choi, Yichi Yang, Yiding Hao, Yifu Chen, Yonatan Belinkov, Yu~Hou, Yufang Hou, Yuntao Bai, Zachary Seid, Zhuoye Zhao, Zijian Wang, Zijie~J. Wang, Zirui Wang, and Ziyi Wu.
  2023.
\newblock \href {https://openreview.net/forum?id=uyTL5Bvosj} {Beyond the imitation game: Quantifying and extrapolating the capabilities of language models}.
\newblock \emph{Transactions on Machine Learning Research}.

\bibitem[{Stanton et~al.(2021)Stanton, Izmailov, Kirichenko, Alemi, and Wilson}]{StantonIKAW21}
Samuel Stanton, Pavel Izmailov, Polina Kirichenko, Alexander~A. Alemi, and Andrew~Gordon Wilson. 2021.
\newblock \href {https://proceedings.neurips.cc/paper/2021/hash/376c6b9ff3bedbbea56751a84fffc10c-Abstract.html} {Does knowledge distillation really work?}
\newblock In \emph{Advances in Neural Information Processing Systems 34: Annual Conference on Neural Information Processing Systems 2021, NeurIPS 2021, December 6-14, 2021, virtual}, pages 6906--6919.

\bibitem[{Sun et~al.(2024)Sun, Liu, Bair, and Kolter}]{sun2024a}
Mingjie Sun, Zhuang Liu, Anna Bair, and J~Zico Kolter. 2024.
\newblock \href {https://openreview.net/forum?id=PxoFut3dWW} {A simple and effective pruning approach for large language models}.
\newblock In \emph{The Twelfth International Conference on Learning Representations}.

\bibitem[{Talmor et~al.(2019)Talmor, Herzig, Lourie, and Berant}]{talmor-etal-2019-commonsenseqa}
Alon Talmor, Jonathan Herzig, Nicholas Lourie, and Jonathan Berant. 2019.
\newblock \href {https://doi.org/10.18653/v1/N19-1421} {{C}ommonsense{QA}: A question answering challenge targeting commonsense knowledge}.
\newblock In \emph{Proceedings of the 2019 Conference of the North {A}merican Chapter of the Association for Computational Linguistics: Human Language Technologies, Volume 1 (Long and Short Papers)}, pages 4149--4158, Minneapolis, Minnesota. Association for Computational Linguistics.

\bibitem[{Tata and Patel(2003)}]{TataP03}
Sandeep Tata and Jignesh~M. Patel. 2003.
\newblock \href {https://doi.org/10.1109/SSDM.2003.1214975} {Piqa: An algebra for querying protein data sets}.
\newblock In \emph{Proceedings of the 15th International Conference on Scientific and Statistical Database Management {(SSDBM} 2003), 9-11 July 2003, Cambridge, MA, {USA}}, pages 141--150. {IEEE} Computer Society.

\bibitem[{Touvron et~al.(2023{\natexlab{a}})Touvron, Lavril, Izacard, Martinet, Lachaux, Lacroix, Rozi{\`{e}}re, Goyal, Hambro, Azhar, Rodriguez, Joulin, Grave, and Lample}]{abs-2302-13971}
Hugo Touvron, Thibaut Lavril, Gautier Izacard, Xavier Martinet, Marie{-}Anne Lachaux, Timoth{\'{e}}e Lacroix, Baptiste Rozi{\`{e}}re, Naman Goyal, Eric Hambro, Faisal Azhar, Aur{\'{e}}lien Rodriguez, Armand Joulin, Edouard Grave, and Guillaume Lample. 2023{\natexlab{a}}.
\newblock \href {https://doi.org/10.48550/ARXIV.2302.13971} {Llama: Open and efficient foundation language models}.
\newblock \emph{CoRR}, abs/2302.13971.

\bibitem[{Touvron et~al.(2023{\natexlab{b}})Touvron, Martin, Stone, Albert, Almahairi, Babaei, Bashlykov, Batra, Bhargava, Bhosale, Bikel, Blecher, Canton{-}Ferrer, Chen, Cucurull, Esiobu, Fernandes, Fu, Fu, Fuller, Gao, Goswami, Goyal, Hartshorn, Hosseini, Hou, Inan, Kardas, Kerkez, Khabsa, Kloumann, Korenev, Koura, Lachaux, Lavril, Lee, Liskovich, Lu, Mao, Martinet, Mihaylov, Mishra, Molybog, Nie, Poulton, Reizenstein, Rungta, Saladi, Schelten, Silva, Smith, Subramanian, Tan, Tang, Taylor, Williams, Kuan, Xu, Yan, Zarov, Zhang, Fan, Kambadur, Narang, Rodriguez, Stojnic, Edunov, and Scialom}]{abs-2307-09288}
Hugo Touvron, Louis Martin, Kevin Stone, Peter Albert, Amjad Almahairi, Yasmine Babaei, Nikolay Bashlykov, Soumya Batra, Prajjwal Bhargava, Shruti Bhosale, Dan Bikel, Lukas Blecher, Cristian Canton{-}Ferrer, Moya Chen, Guillem Cucurull, David Esiobu, Jude Fernandes, Jeremy Fu, Wenyin Fu, Brian Fuller, Cynthia Gao, Vedanuj Goswami, Naman Goyal, Anthony Hartshorn, Saghar Hosseini, Rui Hou, Hakan Inan, Marcin Kardas, Viktor Kerkez, Madian Khabsa, Isabel Kloumann, Artem Korenev, Punit~Singh Koura, Marie{-}Anne Lachaux, Thibaut Lavril, Jenya Lee, Diana Liskovich, Yinghai Lu, Yuning Mao, Xavier Martinet, Todor Mihaylov, Pushkar Mishra, Igor Molybog, Yixin Nie, Andrew Poulton, Jeremy Reizenstein, Rashi Rungta, Kalyan Saladi, Alan Schelten, Ruan Silva, Eric~Michael Smith, Ranjan Subramanian, Xiaoqing~Ellen Tan, Binh Tang, Ross Taylor, Adina Williams, Jian~Xiang Kuan, Puxin Xu, Zheng Yan, Iliyan Zarov, Yuchen Zhang, Angela Fan, Melanie Kambadur, Sharan Narang, Aur{\'{e}}lien Rodriguez, Robert Stojnic, Sergey Edunov,
  and Thomas Scialom. 2023{\natexlab{b}}.
\newblock \href {https://doi.org/10.48550/ARXIV.2307.09288} {Llama 2: Open foundation and fine-tuned chat models}.
\newblock \emph{CoRR}, abs/2307.09288.

\bibitem[{Wang and Komatsuzaki(2021)}]{gpt-j}
Ben Wang and Aran Komatsuzaki. 2021.
\newblock {GPT-J-6B: A 6 Billion Parameter Autoregressive Language Model}.
\newblock \url{https://github.com/kingoflolz/mesh-transformer-jax}.

\bibitem[{Wang et~al.(2023{\natexlab{a}})Wang, Wang, Li, Gao, Yin, and Ren}]{WangWLGYR23}
Peifeng Wang, Zhengyang Wang, Zheng Li, Yifan Gao, Bing Yin, and Xiang Ren. 2023{\natexlab{a}}.
\newblock \href {https://aclanthology.org/2023.acl-long.304} {{SCOTT:} self-consistent chain-of-thought distillation}.
\newblock In \emph{Proceedings of the 61st Annual Meeting of the Association for Computational Linguistics (Volume 1: Long Papers), {ACL} 2023, Toronto, Canada, July 9-14, 2023}, pages 5546--5558. Association for Computational Linguistics.

\bibitem[{Wang et~al.(2023{\natexlab{b}})Wang, Zhu, Saxon, Steyvers, and Wang}]{abs-2301-11916}
Xinyi Wang, Wanrong Zhu, Michael Saxon, Mark Steyvers, and William~Yang Wang. 2023{\natexlab{b}}.
\newblock \href {https://openreview.net/forum?id=BGvkwZEGt7} {Large language models are latent variable models: Explaining and finding good demonstrations for in-context learning}.
\newblock In \emph{Thirty-seventh Conference on Neural Information Processing Systems}.

\bibitem[{Wang et~al.(2023{\natexlab{c}})Wang, Wei, Schuurmans, Le, Chi, Narang, Chowdhery, and Zhou}]{0002WSLCNCZ23}
Xuezhi Wang, Jason Wei, Dale Schuurmans, Quoc~V. Le, Ed~H. Chi, Sharan Narang, Aakanksha Chowdhery, and Denny Zhou. 2023{\natexlab{c}}.
\newblock \href {https://openreview.net/pdf?id=1PL1NIMMrw} {Self-consistency improves chain of thought reasoning in language models}.
\newblock In \emph{The Eleventh International Conference on Learning Representations, {ICLR} 2023, Kigali, Rwanda, May 1-5, 2023}. OpenReview.net.

\bibitem[{Wang et~al.(2023{\natexlab{d}})Wang, Kordi, Mishra, Liu, Smith, Khashabi, and Hajishirzi}]{wang-etal-2023-self-instruct}
Yizhong Wang, Yeganeh Kordi, Swaroop Mishra, Alisa Liu, Noah~A. Smith, Daniel Khashabi, and Hannaneh Hajishirzi. 2023{\natexlab{d}}.
\newblock \href {https://doi.org/10.18653/v1/2023.acl-long.754} {Self-instruct: Aligning language models with self-generated instructions}.
\newblock In \emph{Proceedings of the 61st Annual Meeting of the Association for Computational Linguistics (Volume 1: Long Papers)}, pages 13484--13508, Toronto, Canada. Association for Computational Linguistics.

\bibitem[{Wang et~al.(2023{\natexlab{e}})Wang, Kordi, Mishra, Liu, Smith, Khashabi, and Hajishirzi}]{WangKMLSKH23}
Yizhong Wang, Yeganeh Kordi, Swaroop Mishra, Alisa Liu, Noah~A. Smith, Daniel Khashabi, and Hannaneh Hajishirzi. 2023{\natexlab{e}}.
\newblock \href {https://doi.org/10.18653/v1/2023.acl-long.754} {Self-instruct: Aligning language models with self-generated instructions}.
\newblock In \emph{Proceedings of the 61st Annual Meeting of the Association for Computational Linguistics (Volume 1: Long Papers), {ACL} 2023, Toronto, Canada, July 9-14, 2023}, pages 13484--13508. Association for Computational Linguistics.

\bibitem[{Wang et~al.(2021)Wang, Wang, Joty, and Hoi}]{0034WJH21}
Yue Wang, Weishi Wang, Shafiq~R. Joty, and Steven C.~H. Hoi. 2021.
\newblock \href {https://doi.org/10.18653/V1/2021.EMNLP-MAIN.685} {Codet5: Identifier-aware unified pre-trained encoder-decoder models for code understanding and generation}.
\newblock In \emph{Proceedings of the 2021 Conference on Empirical Methods in Natural Language Processing, {EMNLP} 2021, Virtual Event / Punta Cana, Dominican Republic, 7-11 November, 2021}, pages 8696--8708. Association for Computational Linguistics.

\bibitem[{Wang et~al.(2023{\natexlab{f}})Wang, Huang, Liu, Wang, Song, Zhang, Huang, Wei, Deng, Sun, and Zhang}]{WangHLWSZHWDSZ23}
Zhaoyang Wang, Shaohan Huang, Yuxuan Liu, Jiahai Wang, Minghui Song, Zihan Zhang, Haizhen Huang, Furu Wei, Weiwei Deng, Feng Sun, and Qi~Zhang. 2023{\natexlab{f}}.
\newblock \href {https://aclanthology.org/2023.emnlp-main.120} {Democratizing reasoning ability: Tailored learning from large language model}.
\newblock In \emph{Proceedings of the 2023 Conference on Empirical Methods in Natural Language Processing, {EMNLP} 2023, Singapore, December 6-10, 2023}, pages 1948--1966. Association for Computational Linguistics.

\bibitem[{Wei et~al.(2022)Wei, Wang, Schuurmans, Bosma, Ichter, Xia, Chi, Le, and Zhou}]{Wei0SBIXCLZ22}
Jason Wei, Xuezhi Wang, Dale Schuurmans, Maarten Bosma, Brian Ichter, Fei Xia, Ed~H. Chi, Quoc~V. Le, and Denny Zhou. 2022.
\newblock \href {http://papers.nips.cc/paper\_files/paper/2022/hash/9d5609613524ecf4f15af0f7b31abca4-Abstract-Conference.html} {Chain-of-thought prompting elicits reasoning in large language models}.
\newblock In \emph{NeurIPS}.

\bibitem[{Wei et~al.(2023)Wei, Zhang, Li, Zhang, Gong, Guo, and Liu}]{wei-etal-2023-outlier}
Xiuying Wei, Yunchen Zhang, Yuhang Li, Xiangguo Zhang, Ruihao Gong, Jinyang Guo, and Xianglong Liu. 2023.
\newblock \href {https://doi.org/10.18653/v1/2023.emnlp-main.102} {Outlier suppression+: Accurate quantization of large language models by equivalent and effective shifting and scaling}.
\newblock In \emph{Proceedings of the 2023 Conference on Empirical Methods in Natural Language Processing}, pages 1648--1665, Singapore. Association for Computational Linguistics.

\bibitem[{Wen et~al.(2016)Wen, Wu, Wang, Chen, and Li}]{NIPS2016_41bfd20a}
Wei Wen, Chunpeng Wu, Yandan Wang, Yiran Chen, and Hai Li. 2016.
\newblock \href {https://proceedings.neurips.cc/paper_files/paper/2016/file/41bfd20a38bb1b0bec75acf0845530a7-Paper.pdf} {Learning structured sparsity in deep neural networks}.
\newblock In \emph{Advances in Neural Information Processing Systems}, volume~29. Curran Associates, Inc.

\bibitem[{Williams and Aletras(2023)}]{abs-2311-09755}
Miles Williams and Nikolaos Aletras. 2023.
\newblock \href {https://doi.org/10.48550/ARXIV.2311.09755} {How does calibration data affect the post-training pruning and quantization of large language models?}
\newblock \emph{CoRR}, abs/2311.09755.

\bibitem[{Williams et~al.(2009)Williams, Waterman, and Patterson}]{WilliamsWP09}
Samuel Williams, Andrew Waterman, and David~A. Patterson. 2009.
\newblock \href {https://doi.org/10.1145/1498765.1498785} {Roofline: an insightful visual performance model for multicore architectures}.
\newblock \emph{Commun. {ACM}}, 52(4):65--76.

\bibitem[{Wu et~al.(2024)Wu, Waheed, Zhang, Abdul-Mageed, and Aji}]{wu-etal-2024-lamini}
Minghao Wu, Abdul Waheed, Chiyu Zhang, Muhammad Abdul-Mageed, and Alham Aji. 2024.
\newblock \href {https://aclanthology.org/2024.eacl-long.57} {{L}a{M}ini-{LM}: A diverse herd of distilled models from large-scale instructions}.
\newblock In \emph{Proceedings of the 18th Conference of the European Chapter of the Association for Computational Linguistics (Volume 1: Long Papers)}, pages 944--964, St. Julian{'}s, Malta. Association for Computational Linguistics.

\bibitem[{Xia et~al.(2023)Xia, Zheng, Li, Zhuang, Zhou, Qiu, Li, Lin, and Song}]{10.14778/3626292.3626303}
Haojun Xia, Zhen Zheng, Yuchao Li, Donglin Zhuang, Zhongzhu Zhou, Xiafei Qiu, Yong Li, Wei Lin, and Shuaiwen~Leon Song. 2023.
\newblock \href {https://doi.org/10.14778/3626292.3626303} {Flash-llm: Enabling cost-effective and highly-efficient large generative model inference with unstructured sparsity}.
\newblock \emph{Proc. VLDB Endow.}, 17(2):211–224.

\bibitem[{Xia et~al.(2024)Xia, Gao, Zeng, and Chen}]{xia2024sheared}
Mengzhou Xia, Tianyu Gao, Zhiyuan Zeng, and Danqi Chen. 2024.
\newblock \href {https://openreview.net/forum?id=09iOdaeOzp} {Sheared {LL}a{MA}: Accelerating language model pre-training via structured pruning}.
\newblock In \emph{The Twelfth International Conference on Learning Representations}.

\bibitem[{Xia et~al.(2020)Xia, Wu, and Durme}]{XiaWD20}
Patrick Xia, Shijie Wu, and Benjamin~Van Durme. 2020.
\newblock \href {https://doi.org/10.18653/V1/2020.EMNLP-MAIN.608} {Which *bert? {A} survey organizing contextualized encoders}.
\newblock In \emph{Proceedings of the 2020 Conference on Empirical Methods in Natural Language Processing, {EMNLP} 2020, Online, November 16-20, 2020}, pages 7516--7533. Association for Computational Linguistics.

\bibitem[{Xiao et~al.(2023)Xiao, Lin, Seznec, Wu, Demouth, and Han}]{XiaoLSWDH23}
Guangxuan Xiao, Ji~Lin, Micka{\"{e}}l Seznec, Hao Wu, Julien Demouth, and Song Han. 2023.
\newblock \href {https://proceedings.mlr.press/v202/xiao23c.html} {Smoothquant: Accurate and efficient post-training quantization for large language models}.
\newblock In \emph{International Conference on Machine Learning, {ICML} 2023, 23-29 July 2023, Honolulu, Hawaii, {USA}}, volume 202 of \emph{Proceedings of Machine Learning Research}, pages 38087--38099. {PMLR}.

\bibitem[{Xu et~al.(2021)Xu, Zhou, Ge, Xu, McAuley, and Wei}]{XuZG0MW21}
Canwen Xu, Wangchunshu Zhou, Tao Ge, Ke~Xu, Julian~J. McAuley, and Furu Wei. 2021.
\newblock \href {https://doi.org/10.18653/v1/2021.emnlp-main.832} {Beyond preserved accuracy: Evaluating loyalty and robustness of {BERT} compression}.
\newblock In \emph{Proceedings of the 2021 Conference on Empirical Methods in Natural Language Processing, {EMNLP} 2021, Virtual Event / Punta Cana, Dominican Republic, 7-11 November, 2021}, pages 10653--10659. Association for Computational Linguistics.

\bibitem[{Xu et~al.(2024)Xu, Han, Yang, Wang, Zhu, Liu, Liu, and Che}]{abs-2402-11295}
Yuzhuang Xu, Xu~Han, Zonghan Yang, Shuo Wang, Qingfu Zhu, Zhiyuan Liu, Weidong Liu, and Wanxiang Che. 2024.
\newblock \href {https://doi.org/10.48550/ARXIV.2402.11295} {Onebit: Towards extremely low-bit large language models}.
\newblock \emph{CoRR}, abs/2402.11295.

\bibitem[{Yao et~al.(2022)Yao, Aminabadi, Zhang, Wu, Li, and He}]{YaoAZWLH22}
Zhewei Yao, Reza~Yazdani Aminabadi, Minjia Zhang, Xiaoxia Wu, Conglong Li, and Yuxiong He. 2022.
\newblock \href {http://papers.nips.cc/paper\_files/paper/2022/hash/adf7fa39d65e2983d724ff7da57f00ac-Abstract-Conference.html} {Zeroquant: Efficient and affordable post-training quantization for large-scale transformers}.
\newblock In \emph{NeurIPS}.

\bibitem[{Yao et~al.(2023)Yao, Li, Wu, Youn, and He}]{abs-2303-08302}
Zhewei Yao, Cheng Li, Xiaoxia Wu, Stephen Youn, and Yuxiong He. 2023.
\newblock \href {https://doi.org/10.48550/arXiv.2303.08302} {Zeroquant-v2: Exploring post-training quantization in llms from comprehensive study to low rank compensation}.
\newblock \emph{CoRR}, abs/2303.08302.

\bibitem[{Yuan et~al.(2023{\natexlab{a}})Yuan, Niu, Liu, Liu, Wang, Shang, Sun, Wu, Wu, and Wu}]{abs-2304-01089}
Zhihang Yuan, Lin Niu, Jiawei Liu, Wenyu Liu, Xinggang Wang, Yuzhang Shang, Guangyu Sun, Qiang Wu, Jiaxiang Wu, and Bingzhe Wu. 2023{\natexlab{a}}.
\newblock \href {https://doi.org/10.48550/arXiv.2304.01089} {{RPTQ:} reorder-based post-training quantization for large language models}.
\newblock \emph{CoRR}, abs/2304.01089.

\bibitem[{Yuan et~al.(2023{\natexlab{b}})Yuan, Shang, Song, Wu, Yan, and Sun}]{abs-2312-05821}
Zhihang Yuan, Yuzhang Shang, Yue Song, Qiang Wu, Yan Yan, and Guangyu Sun. 2023{\natexlab{b}}.
\newblock \href {https://doi.org/10.48550/ARXIV.2312.05821} {{ASVD:} activation-aware singular value decomposition for compressing large language models}.
\newblock \emph{CoRR}, abs/2312.05821.

\bibitem[{Yue et~al.(2024)Yue, Yuan, Duanmu, Zhou, Wu, and Nie}]{abs-2402-12065}
Yuxuan Yue, Zhihang Yuan, Haojie Duanmu, Sifan Zhou, Jianlong Wu, and Liqiang Nie. 2024.
\newblock \href {https://doi.org/10.48550/ARXIV.2402.12065} {Wkvquant: Quantizing weight and key/value cache for large language models gains more}.
\newblock \emph{CoRR}, abs/2402.12065.

\bibitem[{Zhang et~al.(2022)Zhang, Roller, Goyal, Artetxe, Chen, Chen, Dewan, Diab, Li, Lin, Mihaylov, Ott, Shleifer, Shuster, Simig, Koura, Sridhar, Wang, and Zettlemoyer}]{abs-2205-01068}
Susan Zhang, Stephen Roller, Naman Goyal, Mikel Artetxe, Moya Chen, Shuohui Chen, Christopher Dewan, Mona~T. Diab, Xian Li, Xi~Victoria Lin, Todor Mihaylov, Myle Ott, Sam Shleifer, Kurt Shuster, Daniel Simig, Punit~Singh Koura, Anjali Sridhar, Tianlu Wang, and Luke Zettlemoyer. 2022.
\newblock \href {https://doi.org/10.48550/ARXIV.2205.01068} {{OPT:} open pre-trained transformer language models}.
\newblock \emph{CoRR}, abs/2205.01068.

\bibitem[{Zhang et~al.(2024)Zhang, Zhao, Lin, Yunyun, Yao, Han, Tanner, Liu, and Ji}]{zhang2024dynamic}
Yuxin Zhang, Lirui Zhao, Mingbao Lin, Sun Yunyun, Yiwu Yao, Xingjia Han, Jared Tanner, Shiwei Liu, and Rongrong Ji. 2024.
\newblock \href {https://openreview.net/forum?id=1ndDmZdT4g} {Dynamic sparse no training: Training-free fine-tuning for sparse {LLM}s}.
\newblock In \emph{The Twelfth International Conference on Learning Representations}.

\bibitem[{Zhao et~al.(2023)Zhao, Zhou, Li, Tang, Wang, Hou, Min, Zhang, Zhang, Dong, Du, Yang, Chen, Chen, Jiang, Ren, Li, Tang, Liu, Liu, Nie, and Wen}]{abs-2303-18223}
Wayne~Xin Zhao, Kun Zhou, Junyi Li, Tianyi Tang, Xiaolei Wang, Yupeng Hou, Yingqian Min, Beichen Zhang, Junjie Zhang, Zican Dong, Yifan Du, Chen Yang, Yushuo Chen, Zhipeng Chen, Jinhao Jiang, Ruiyang Ren, Yifan Li, Xinyu Tang, Zikang Liu, Peiyu Liu, Jian{-}Yun Nie, and Ji{-}Rong Wen. 2023.
\newblock \href {https://doi.org/10.48550/arXiv.2303.18223} {A survey of large language models}.
\newblock \emph{CoRR}, abs/2303.18223.

\bibitem[{Zheng et~al.(2023)Zheng, Chiang, Sheng, Zhuang, Wu, Zhuang, Lin, Li, Li, Xing, Zhang, Gonzalez, and Stoica}]{vicuna2023}
Lianmin Zheng, Wei{-}Lin Chiang, Ying Sheng, Siyuan Zhuang, Zhanghao Wu, Yonghao Zhuang, Zi~Lin, Zhuohan Li, Dacheng Li, Eric~P. Xing, Hao Zhang, Joseph~E. Gonzalez, and Ion Stoica. 2023.
\newblock \href {http://papers.nips.cc/paper\_files/paper/2023/hash/91f18a1287b398d378ef22505bf41832-Abstract-Datasets\_and\_Benchmarks.html} {Judging llm-as-a-judge with mt-bench and chatbot arena}.
\newblock In \emph{Advances in Neural Information Processing Systems 36: Annual Conference on Neural Information Processing Systems 2023, NeurIPS 2023, New Orleans, LA, USA, December 10 - 16, 2023}.

\bibitem[{Zhu et~al.(2024)Zhu, Qi, Zhang, Long, Lin, and Zhou}]{abs-2305-13888}
Xuekai Zhu, Biqing Qi, Kaiyan Zhang, Xinwei Long, Zhouhan Lin, and Bowen Zhou. 2024.
\newblock \href {https://aclanthology.org/2024.naacl-long.142} {{P}a{D}: Program-aided distillation can teach small models reasoning better than chain-of-thought fine-tuning}.
\newblock In \emph{Proceedings of the 2024 Conference of the North American Chapter of the Association for Computational Linguistics: Human Language Technologies (Volume 1: Long Papers)}, pages 2571--2597, Mexico City, Mexico. Association for Computational Linguistics.

\bibitem[{Zoph and Le(2017)}]{ZophL17}
Barret Zoph and Quoc~V. Le. 2017.
\newblock \href {https://openreview.net/forum?id=r1Ue8Hcxg} {Neural architecture search with reinforcement learning}.
\newblock In \emph{5th International Conference on Learning Representations, {ICLR} 2017, Toulon, France, April 24-26, 2017, Conference Track Proceedings}. OpenReview.net.

\end{thebibliography}
\bibliographystyle{acl_natbib}

\end{document}